\documentclass{article}

\usepackage{arxiv}

\usepackage[utf8]{inputenc} 
\usepackage[T1]{fontenc}    
\usepackage{url}            
\usepackage{amsfonts}       
\usepackage{nicefrac}       
\usepackage{microtype}      
\usepackage{doi}
\usepackage{graphicx} 
\usepackage{booktabs} 
\usepackage[round]{natbib}

\usepackage{amsthm}         
\usepackage{siunitx}        
\usepackage{threeparttable} 
\usepackage{caption}        
\usepackage{algorithm}
\usepackage{algorithmic}
\usepackage{makecell}

\usepackage{newfloat}
\usepackage{listings}
\DeclareCaptionStyle{ruled}{labelfont=normalfont,labelsep=colon,strut=off} 
\floatstyle{ruled}
\newfloat{listing}{tb}{lst}{}
\floatname{listing}{Listing}
\title{\textbf{MEGA-GUI: Multi-stage Enhanced Grounding Agents for GUI Elements}}

\author{
    SeokJoo Kwak \\
    Samsung SDS \\
    \texttt{seokjoo.kwak@samsung.com}
    \And
    Jihoon Kim \\
    Samsung SDS \\
    \texttt{jihoon00.kim@samsung.com}
    \And
    Boyoun Kim \\
    Samsung SDS \\
    \texttt{boyoun1.kim@samsung.com}
    \And
    Jung Jae Yoon \\
    Samsung SDS \\
    \texttt{alton.jung@samsung.com}
    \And
    Wooseok Jang \\
    Samsung SDS \\
    \texttt{wseok.jang@samsung.com}
    \And
    Jeonghoon Hong \\
    Samsung SDS \\
    \texttt{jh74.hong@samsung.com}
    \And
    Jaeho Yang \\
    Samsung SDS \\
    \texttt{jaeho3.yang@samsung.com}
    \And
    Yeong-Dae Kwon \\
    Samsung SDS \\
    \texttt{y.d.kwon@samsung.com}
}

\date{}

\renewcommand{\shorttitle}{MEGA-GUI: Multi-stage Enhanced Grounding Agents for GUI Elements}

\hypersetup{
    colorlinks=true,
    linkcolor=blue,
    filecolor=magenta,      
    urlcolor=cyan,
    citecolor=blue,
    pdftitle={MEGA-GUI: Multi-stage Enhanced Grounding Agents for GUI Elements},
    pdfauthor={SeokJoo Kwak, et al.},
}

\begin{document}

\maketitle

\begin{abstract}
Graphical User Interface (GUI) grounding---the task of mapping natural language instructions to screen coordinates---is critical for autonomous agents and accessibility technologies. Prior methods, often built on monolithic models or one-shot, unidirectional pipelines, lack the modularity and error recovery needed to handle visual clutter and ambiguous instructions. We introduce \textbf{MEGA-GUI}, a multi-stage framework that decomposes GUI grounding into coarse Region-of-Interest (ROI) selection and fine-grained element grounding, orchestrating specialized agents based on diverse Vision-Language Models (VLMs). Central to this framework is a bidirectional ROI zoom algorithm that alleviates spatial dilution, complemented by a context-aware rewriting agent that reduces semantic ambiguity. Through analysis, we show that VLMs possess complementary strengths and weaknesses depending on the input’s visual scale and that leveraging these differences with a modular design substantially outperforms any monolithic approach.  On the visually dense \textit{ScreenSpot-Pro} benchmark, MEGA-GUI achieves 73.18\% accuracy; on the semantically complex \textit{OSWorld-G} benchmark, it scores 68.63\%, significantly surpassing prior results. To support reproducibility and future research, we release all code and the Grounding Benchmark Toolkit (GBT) at: \url{https://github.com/samsungsds-research-papers/mega-gui}.
\end{abstract}


\section{INTRODUCTION}

Recent breakthroughs in multimodal AI, exemplified by agents leveraging OpenAI's Computer Use API with GPT-4o \citep{openai_computer_use_2024} and Anthropic's screen-aware Claude models \citep{anthropic_computer_use_2024}, have opened a new frontier in autonomous computer control. These systems promise to revolutionize domains ranging from enterprise software automation to personal assistive technologies. However, their practical effectiveness remains fundamentally limited by a persistent and critical challenge: GUI grounding---the task of accurately mapping natural language commands to specific screen coordinates. Despite significant advances, current agents still exhibit high error rates on complex grounding benchmarks, a critical limitation documented across numerous studies \citep{wu2025gui, li2025screenspot, xie2025scalingcomputerusegroundinguser} rendering them unreliable for most real-world scenarios.

This difficulty stems from a dual challenge. First, the proliferation of high-resolution displays leads to extreme spatial dilution, creating a ``needle-in-a-haystack'' problem where target widgets may occupy less than 0.1\% of the total screen area \citep{wu2025gui}. This saturates a model's perceptual capacity with distractors, making accurate localization exceedingly difficult. Second, user instructions are often semantically ambiguous. Commands like ``click the small wrench icon'' or ``close the dialog'' require contextual reasoning that goes beyond simple visual pattern matching \citep{liu2025ui}. Failures to address these intertwined issues in cluttered, real-world environments underscore the urgent need for grounding methodologies that are both resilient and precise.

To address this, the dominant approach has been to train large, monolithic Vision-Language Models (VLMs) on massive, annotated GUI datasets \citep{you2024ferretuigroundedmobileui, chen2025guicoursegeneralvisionlanguage}. While these models have pushed the field forward, they treat grounding as a single, end-to-end task, which can limit their adaptability. This limitation mirrors a broader paradigm shift within the AI community; the research focus is evolving from enhancing the performance of monolithic models on single-query tasks towards designing and orchestrating multi-agent systems to solve complex, multi-step problems \citep{wang2024survey}. This evolution presents a clear opportunity to reframe GUI grounding---not as a monolithic prediction problem, but as a structured control task.

Building on this, a second strategy explores agentic, inference-time refinement pipelines that iteratively narrow the region of interest \citep{luo2025visual, li2025screenspot}. While this marks progress toward agentic control, these methods remain fundamentally fragile: their unidirectional refinement paths lack mechanisms for backtracking or context recovery, so early errors cannot be corrected. In addition, such frameworks are often assembled in an ad-hoc fashion, hard-wiring a single VLM into a fixed pipeline and overlooking a crucial insight: the sub-tasks of grounding---broad contextual search and fine-grained element identification---differ substantially in nature and place very different demands on the system.

In this paper, we introduce \textbf{MEGA-GUI}, a modular framework that reformulates GUI element grounding as a structured, multi-stage control problem. By orchestrating powerful VLMs as specialized ``tools'' within a stateful graph, MEGA-GUI achieves new levels of robustness and precision. Our work contributes a novel methodology specifically tailored to the GUI grounding task, moving beyond ad-hoc system integration toward principled design.

Its primary contributions are threefold:
\begin{enumerate}
    \item \textbf{A Principled, Decoupled Framework:} We propose a novel architecture that separates coarse spatial localization from fine-grained semantic grounding. Its centerpiece is a robust \textbf{bidirectional adaptive zoom} algorithm, a recoverable search mechanism that uses geometric feedback to proactively refine the agent's visual focus and recover from initial prediction errors.

    \item \textbf{Empirical Validation of a ``No Free Lunch'' Principle:} MEGA-GUI establishes new state-of-the-art performance on the ScreenSpot-Pro and OSWorld-G benchmarks. We provide the first empirical validation that a ``No Free Lunch'' principle applies to VLM specialization, grounded in a systematic study of model performance across a spectrum of visual scales (window sizes). Our analysis reveals that different models possess distinct, scale-dependent strengths for coarse visual search versus precise, fine-grained identification, providing a principled, data-driven basis for our modular agent design.
        
    \item \textbf{An Open-Source Toolkit for Principled Design:} We release our \textbf{Grounding Benchmark Toolkit (GBT)}, a comprehensive suite for the systematic evaluation of GUI grounding components. By enabling a data-driven methodology for agent design, the GBT fosters more reproducible, transparent, and rigorous research, aligning with the community's growing emphasis on responsible AI development.
\end{enumerate}

\section{RELATED WORK}
Early GUI automation relied on structured metadata like accessibility trees~\citep{Li_2021, tang2025thinktwiceclickonce, fan2025guibeealignguiaction}, but the prevalence of custom-rendered widgets in modern applications has necessitated a shift toward vision-based agents that operate directly on raw pixels. These vision-centric agents have evolved along several primary architectural paradigms, each making significant contributions to the field.

\textbf{General Purpose Vision-Language Models.} Pre-trained VLMs such as CUA~\citep{OPENAICUA}, Gemini~\citep{team2023gemini}, and Qwen~\citep{qwen2025qwen25technicalreport} represent a foundational pillar in this area. Trained on vast datasets of image-text pairs~\citep{dosovitskiy2020image}, these models possess remarkable general visual understanding, enabling them to interpret a wide array of GUI elements. Their versatility provides a powerful starting point for numerous applications, including interface navigation and user intent understanding, establishing a robust baseline upon which more specialized systems can be built.

\paragraph{Specialized Models for GUI Grounding.}
Building upon this foundation, a key research direction involves creating specialized agents through fine-tuning, with a clear evolution in both the data and the methods used for training. One line of work focuses on the data source: models like SeeClick~\citep{cheng2024seeclick} and UGround~\citep{gou2024navigating} leverage scalable but sometimes noisy web data , while UI-TARS~\citep{qin2025ui} uses high-precision manual annotations, and JEDI~\citep{xie2025scalingcomputerusegroundinguser} pioneers large-scale data synthesis to combine the benefits of both. A parallel evolution is occurring in training methodology. While most early models relied on Supervised Fine-Tuning (SFT), recent models like GTA1~\citep{yang2025gta1} have introduced reinforcement learning (RL) to the grounding task. This approach moves beyond predicting a single center point by rewarding any correct click within a target element, better aligning the training objective with the real-world success condition.

\textbf{Integrated Refinement Pipelines.} A third paradigm focuses on enhancing the zero-shot performance of VLMs through innovative inference-time strategies. These methods have pioneered the use of iterative refinement of a Region of Interest (ROI) to improve grounding precision. For instance, ScreenSeekeR~\citep{li2025screenspot} leverages GUI knowledge to guide a cascaded search, while RegionFocus~\citep{luo2025visual} dynamically adjusts focus in response to agent errors. Similarly, DiMo-GUI~\citep{wu2025dimo} employs a modality-aware pipeline with an iterative zoom-in strategy, and ReGUIDE~\citep{lee2025reguide} integrates spatial search with coordinate aggregation. While these approaches excel in cases where initial predictions are near targets, their unidirectional search paths are fundamentally brittle; an early error can propagate irreversibly, especially in dense or ambiguous UIs. MEGA-GUI directly addresses this core limitation by introducing a bidirectional zoom algorithm, providing a robust error recovery mechanism that these unidirectional pipelines lack.

\section{THE MEGA-GUI FRAMEWORK}
\begin{figure}[t]
\centering
\includegraphics[width=.99\columnwidth]{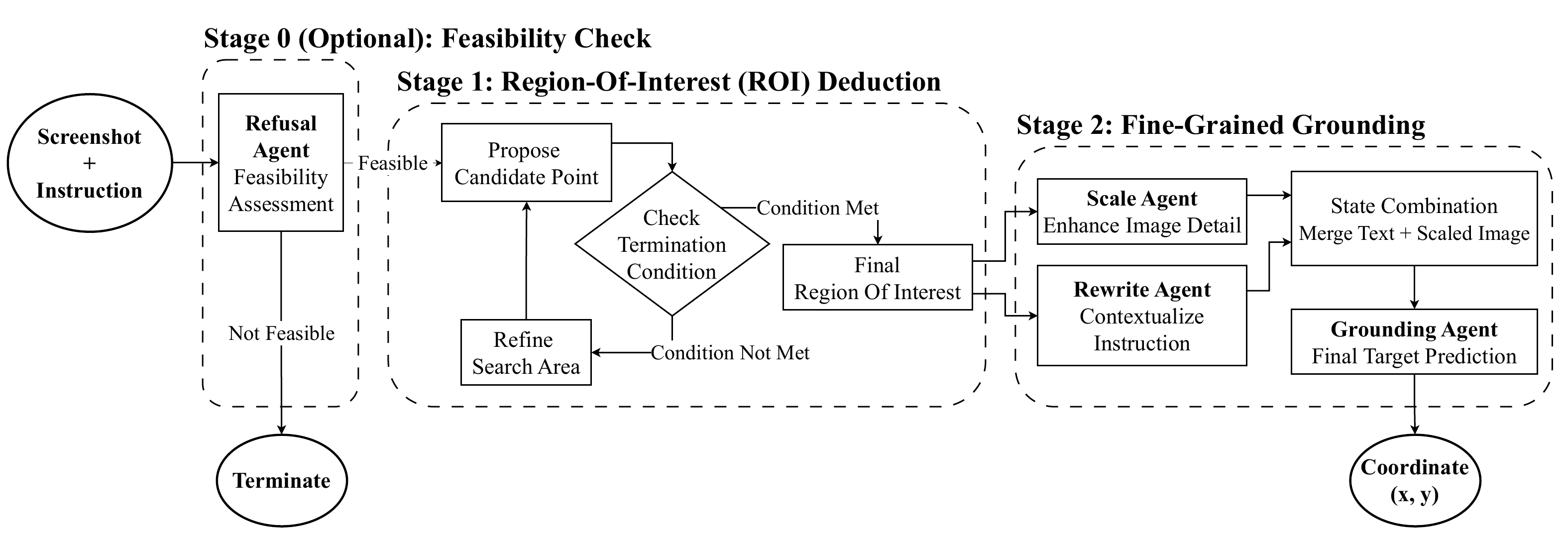}
\caption{The MEGA-GUI Framework. The GUI grounding task is decomposed into three independent stages, each with its own objective. This design enables the modular composition of specialized agents and facilitates systematic evaluation using our Grounding Benchmark Toolkit.}
\label{fig:system_diagram}
\end{figure}

All Vision-Language Models (VLMs)---regardless of whether they are fine-tuned for grounding tasks---struggle to accurately localize target GUI elements in a single pass (as shown in Table~\ref{tab:main_results}). This limitation becomes particularly evident when processing large, high-resolution screen images, often rendering such models unreliable for real-world automation tasks that operate on a user's full-screen interface. To address this challenge, we adopt a divide-and-conquer strategy that breaks the problem into smaller, more manageable subtasks. Our MEGA-GUI framework, illustrated in Figure~\ref{fig:system_diagram}, reformulates GUI grounding as a structured, multi-stage problem-solving process. Each stage is designed with independent objectives, enabling modular evaluation and targeted optimization.

\textbf{Stage 0 (Optional): Feasibility Check.} This stage determines whether the user’s instruction is executable on the current UI screen. An effective implementation should exhibit high \textit{refusal accuracy}---the ability to correctly identify infeasible instructions---while maintaining a low \textit{false positive rate}.

\textbf{Stage 1: Region-of-Interest (ROI) Deduction.} This stage selects a fixed-size ROI that is small enough for a Vision-Language Model (VLM) to perform accurate grounding in the next step. Performance is measured by the \textit{containment rate}, which indicates how often the target GUI element falls within the selected ROI.

\textbf{Stage 2: Fine-Grained Grounding.} Given the ROI, this stage pinpoints the precise coordinates of the target element. The key metric is \textit{accuracy}---the rate of correctly identifying the target, assuming the ROI includes it.

Each stage can be evaluated independently, enabling targeted improvements without introducing interdependencies. This modular design allows for flexible tuning and testing, which in turn enhances overall performance. Notably, the framework is model-agnostic, and we are not restricted to any single VLM. As we will demonstrate, different VLMs---whether open-source or commercial---exhibit complementary strengths. By assigning the most suitable VLM to each stage, MEGA-GUI leverages this heterogeneity to achieve superior results.

\subsection{Bidirectional ROI Zoom}

At the core of MEGA-GUI is the \textbf{Bidirectional ROI Zoom agent}, a novel, closed-loop control algorithm designed for robust visual search and ROI refinement. The core logic is presented in Algorithm~\ref{alg:stable_search_zoom}, and the complete algorithm is detailed in the Appendix. Starting from the full-screen image, the agent progressively reduces the ROI size to a scale at which VLMs perform most effectively. This design addresses a critical failure mode of conventional unidirectional zooming methods, which often become permanently ``stuck'' after an incorrect prediction. By integrating a self-correcting recovery mechanism, our agent can dynamically re-center and resume its search, even after errors.

\textbf{Zoom-In:} When the VLM’s predicted location $p$ for the target element falls within the current ROI, the agent interprets this as a reliable signal. The ROI is then aggressively cropped inward from the boundary farthest from $p$, using a reduction factor $\Delta_{in}$, to refine the search space.

\textbf{Zoom-Out \& Correction:} If the prediction $p$ lies outside the ROI, indicating that the model is unable to identify a feasible target within the current view, the agent uniformly expands the ROI by a factor of $\Delta_{out}$ to restore broader context. To prevent uncontrolled expansion, a cumulative error counter $E$ is maintained. We use a cumulative count, rather than resetting on success, to penalize agents that oscillate between correct and incorrect regions, ensuring a more stable convergence path. If the counter exceeds a predefined threshold $E_{max}$, a corrective zoom-in is triggered instead of further zooming out.

\textbf{Termination:} The refinement process continues until one of two termination conditions is met: (1) \textit{Minimum Area Threshold}, where the ROI's area shrinks below a predefined target size; or (2) \textit{Convergence}, where at least $N_{stable}$ consecutive, in-bounds predictions $p$ are tightly clustered within a localized neighborhood. The convergence condition allows for early termination, avoiding redundant zoom-in operations when the model demonstrates stable and confident predictions.

\begin{algorithm}[tb]
\caption{Bidirectional ROI Zoom (abstracted ver.)}
\label{alg:stable_search_zoom}
\begin{algorithmic}[1] 
\REQUIRE Instruction $q$, Image $I$, Model $\mathcal{M}$
\REQUIRE Hyperparameters: $\Delta_{in}, \Delta_{out}, E_{max}, S_{min}, N_{stable}, \epsilon_{stable}$
\COMMENT{$\Delta$: zoom factors, $E_{max}$: max errors, $S_{min}$: min area}
\COMMENT{$N_{stable}$: Number of points to check for stability (e.g., 3)}
\COMMENT{$\epsilon_{stable}$: max distance for points to be considered stable}
\ENSURE Final cropped Region of Interest

\STATE $\text{ROI} \leftarrow \text{GetImageBounds}(I)$
\STATE $H \leftarrow \emptyset$ \COMMENT{Ordered list of recent in-bounds predictions}
\STATE $E \leftarrow 0$ \COMMENT{Cumulative out-of-bounds error count}
\STATE \COMMENT{$p$ will store the predicted coordinate $(x,y)$}

\WHILE{GetArea(ROI) $>$ $S_{min}$ \textbf{and} \\ \hspace{1em} not SearchConverged($H, N_{stable}, \epsilon_{stable}$)}
    \STATE $p \leftarrow \mathcal{M}.\text{predict}(q, \text{ROI})$
    
    \IF{$p \in \text{ROI}$}
        \STATE $H \leftarrow H \cup \{p\}$
        \STATE $\text{ROI} \leftarrow \text{ZoomIn}(\text{ROI}, p, \Delta_{in})$
    \ELSE
        \STATE $E \leftarrow E + 1$
        \IF{$E \ge E_{max}$}
            \STATE $\text{ROI} \leftarrow \text{ZoomIn}(\text{ROI}, \Delta_{in})$
        \ELSE
            \STATE $\text{ROI} \leftarrow \text{ZoomOut}(\text{ROI}, \Delta_{out})$
        \ENDIF
    \ENDIF
\ENDWHILE

\STATE \textbf{return} $\text{Finalize}(\text{ROI}, H)$
\end{algorithmic}
\end{algorithm}

\subsection{Specialized Agents for Precision Grounding}

\paragraph{The Context-Aware Rewrite Agent.} This agent resolves semantic ambiguity by transforming vague user instructions into precise, context-rich commands. For example, a generic command like \texttt{``Share this''} within a complex application like Figma is ambiguous. The Rewrite Agent transforms it into an explicit directive, such as\texttt{``Click the main `Share' button, the blue button in the top-right toolbar for Figma''}. This process is crucial for eliminating guesswork and preventing errors.

The agent's workflow is deliberately constrained to the focused UI crop provided by the Bidirectional Zoom Agent, forcing it to reason based on the most relevant visual evidence. It operates in two stages: first, it infers the active application's name (e.g., \texttt{app\_name="Figma"}). from the visual cues within the crop. Second, it synthesizes this application context, the user's raw instruction, and the cropped image into a single prompt for a powerful vision-language model like GPT-4o. By analyzing the local text labels, icon colors, and spatial relationships through the lens of this rich context, the agent generates a new, unambiguous instruction. This contextual enrichment significantly reduces the cognitive load on the downstream Grounding Agent and minimizes errors arising from ambiguous commands.

\paragraph{The Conservative Scale Agent}
This agent serves as a deterministic accuracy enhancer. Once the Zoom Agent identifies a focused ROI, the Scale Agent upscales the cropped image back to high resolution before grounding. We have empirically found that this improves the performance of grounding VLMs (see Table~\ref{tab:ablation}; detailed analysis in the Appendix). This approach leverages the VLMs’ capacity to process high-resolution inputs while benefiting from the reduced visual clutter of the cropped ROI, allowing the model to focus its full computational power on the most salient elements.

\section{EXPERIMENTS AND ANALYSIS}
We conduct a rigorous set of experiments to validate the effectiveness of the MEGA-GUI framework, carefully analyzing and optimizing the contribution of each individual component. All experiments are fully reproducible using our publicly available open-source code and the test sets included in GBT.

\subsection{Experimental Setup}

\textbf{Benchmarks.} We evaluate MEGA-GUI on two GUI grounding benchmarks: \textbf{ScreenSpot-Pro (SSP)}~\citep{li2025screenspot}, consisting of 1,581 tasks on 4K-resolution professional software interfaces, designed to test spatial precision under extreme visual density; and \textbf{OSWorld-G (OSG)}~\citep{xie:25}, containing 564 tasks on standard 1080p operating system interfaces, intended to assess functional commonsense in everyday GUI environments. Our primary metric is Top-1 Accuracy where a prediction is correct if its coordinate falls within the ground-truth element's annotated region. This region is a bounding box for the ScreenSpot-Pro dataset and a more precise bounding polygon for the OSWorld-G dataset. 

\textbf{Models and Baselines.} Our framework utilizes powerful, proprietary models for various agentic roles, including localization, grounding, and refinement. These models were accessed via their official APIs during our main experimental period (July-August 2025). They include OpenAI's GPT-4o (model version \texttt{gpt-4o-1120}) and Google's Gemini-2.5-Pro. All open-source models used for comparisons were run locally on our server equipped with eight NVIDIA V100 GPUs. For a comprehensive list of implementation details, please refer to the Appendix.

To rigorously evaluate MEGA-GUI, we compare it against two categories of baselines. For state-of-the-art \textbf{Agentic Methods} (e.g., RegionFocus, ScreenSeekeR), we report the performance figures as published in their respective original papers. For \textbf{Single-Shot VLMs}, we reproduced the experiments to ensure a fair and controlled comparison. This category includes both open-source models (e.g., UI-TARS, Qwen-VL, GTA1) run locally, and proprietary API-based models evaluated as single-shot baselines. A key proprietary baseline is OpenAI's Computer Use Agent (CUA), for which we utilized the \texttt{computer-use-preview} model (v. \texttt{2025-03-11}) via the Azure OpenAI Service. This unified evaluation approach was necessary to mitigate potential variances arising from differing hardware or the absence of publicly available, optimized inference prompts from the original works.

\subsection{Main Results and Component Contributions}

As reported in Table~\ref{tab:main_results}, our optimal MEGA-GUI configuration establishes a new state of the art, achieving \textbf{73.18\%} accuracy on SSP and \textbf{68.63\%} on OSG, significantly surpassing previously published results. These gains are realized through a synergistic composition of specialized agents: Gemini 2.5 Pro powers the adaptive ROI zoom algorithm towards target ROI size of 1000 pixels, GPT-4o performs context-aware instruction rewriting, and UI-TARS-72B carries out the final precision grounding on a $3\times$ bicubic-upscaled image crop. This configuration substantiates the central premise of MEGA-GUI: that an agentic architecture, in which specialized modules are orchestrated to tackle distinct sub-tasks, can achieve superior robustness and accuracy.

\begin{table}[t]
\centering
\begin{threeparttable}
\caption{MEGA-GUI performance on the SSP and OSG, compared against single-shot and agentic baseline methods.}
\label{tab:main_results}

\sisetup{
 table-align-text-post=false,
 detect-weight,
 mode=text,
 table-space-text-post = N/A
}
\centering
\begin{tabular}{
 l
 S[table-format=2.2]
 S[table-format=2.2]
}
\toprule
\textbf{Method} & {\textbf{SSP (\%)}} & {\textbf{OSG (\%)}} \\
\midrule

\multicolumn{3}{l}{\textit{Single-Shot Baselines (Reproduced)}} \\
Gemini-2.5-Pro    & 6.96  & 28.19 \\
CUA (Operator)    & 35.03 & 32.26 \\
UI-TARS-7B        & 29.53 & 44.50 \\
Qwen-VL-2.5-72B   & 41.55 & 46.80 \\
GTA1-72B          & 46.34 & 50.53 \\
Jedi-7B           & 30.10 & 51.41 \\
UI-TARS-72B       & 37.12 & 55.67 \\
GTA1-7B           & 50.03 & 56.22 \\
\midrule

\multicolumn{3}{l}{\textit{Agentic Baselines (Reported in Original Papers)}\tnote{*}} \\
ReGUIDE-7B        & 44.40 & {N/A} \\
ScreenSeekeR      & 48.10 & {N/A} \\
DiMo-GUI          & 49.70 & {N/A} \\
RegionFocus       & 61.60 & {N/A} \\
\midrule

\textbf{MEGA-GUI (Ours)} & \bfseries 73.18 & \bfseries 68.63 \\

\bottomrule
\end{tabular}
\begin{tablenotes}
    \item[*]  {\textit{Note: Performance figures for agentic baselines are as reported in their original publications. To ensure a controlled comparison, all single-shot baselines were reproduced in our unified experimental environment.}}
\end{tablenotes}
\end{threeparttable}
\end{table}

To isolate the contribution of each framework component, we conducted a sequential ablation study, reporting marginal gains over a UI-TARS-72B baseline in Table~\ref{tab:ablation}. The results show that the optimal agent configuration is closely tied to the nature of the benchmark. On the spatially dense SSP benchmark, the ROI zoom agent yields the largest improvement (+28.47,pp), directly addressing the ``needle-in-a-haystack'' challenge. In contrast, on the semantically complex OSG benchmark, the instruction-rewrite agent is most decisive, producing the largest single gain (+9.61,pp) by resolving command ambiguity. In settings where context dominates, the advantage of sophisticated search diminishes, while the importance of semantic disambiguation becomes more pronounced.\footnote{This trend is further confirmed on ScreenSpot v2 (SPv2), a lower-resolution, context-dominant benchmark. Starting from a 90.30\% baseline, the sequential deltas were: ROI Zoom $-0.91$,pp (to 89.39\%), Image Scaling $+0.58$,pp (to 89.97\%), and Instruction Rewrite $+3.35$,pp (to 93.32\%), for a net gain of $+3.02$,pp.}

\begin{table}[t]
\centering
\begin{threeparttable}
\caption{Marginal performance gains (in pp) for each component added sequentially over a UI-TARS-72B baseline.}
\label{tab:ablation}

\sisetup{
 table-align-text-post=false, 
 detect-weight, 
 mode=text, 
 table-space-text-post = N/A,
 table-format=+2.2
}
\centering
\begin{tabular}{
 l
 S
 S
}
\toprule
\textbf{Component Added} & {\textbf{$\Delta$ SSP (pp)}} & {\textbf{$\Delta$ OSG (pp)}} \\
\midrule
+ ROI Zoom (1000 pixel) & {+28.47} & {+2.17} \\
+ Image Scaling (3$\times$) & {+2.40} & {+1.18} \\
+ Instruction Rewrite & {+5.19} & {+9.61}\\
\midrule
\textbf{Total Gain over Baseline} & \textbf{+36.06} & \textbf{+12.96} \\
\bottomrule
\end{tabular}
\begin{tablenotes}
    \item[*] \textit{Note: the single-shot grounding baselines by UI-TARS-72B are 37.12\% on SSP and 55.67\% on OSG.}
\end{tablenotes}
\end{threeparttable}
\end{table}

\paragraph{Impact of Bidirectional Zoom.}
We ablate the zoom policy by comparing our full method to a unidirectional ``zoom-in only'' variant. On ScreenSpot-Pro, across twelve model and hyperparameter configurations, the bidirectional policy yields consistent absolute accuracy gains of $+1.14$--$+4.36$\,pp (mean $+2.22$\,pp). These improvements are stable across settings, indicating that bidirectional zoom is a primary driver of robustness rather than a cosmetic tweak: the ability to zoom out and re-center provides explicit error recovery when early predictions drift, which a purely unidirectional refinement cannot offer.

\paragraph{Validating the Asymmetric Zoom-In Heuristic.}
While iterative refinement methods often use a simple symmetric crop to narrow the search space \citep{li2025screenspot}, we propose a more principled asymmetric zoom-in heuristic. To quantify its contribution, we conducted an ablation study comparing our method against the symmetric baseline. Our approach operates on the principle that a confident VLM prediction identifies the semantic centroid of the target region, which provides a reliable signal for the target's location; therefore, the most efficient search space reduction is to prune the region farthest from this signal. On the ScreenSpot-Pro benchmark, this principled approach proved significantly more effective. Our asymmetric method achieved an ROI containment rate of 89.69\%, outperforming the symmetric baseline's 81.65\% by a margin of 8.04 pp. This result demonstrates that preserving contextual information near the prediction, while aggressively removing distant and likely irrelevant distractors, is a more robust strategy for maintaining target containment during the refinement process.

\subsection{Experiment Details}

\paragraph{Feasibility Check}

We implemented the Refusal Agent using Gemini 2.5 Pro, which takes the user instruction and screen image as input to detect infeasible requests. We envision that this agent can be further enhanced to handle overly ambiguous instructions by prompting the user for clarification, or to identify potentially unsafe commands and halt execution to ensure safety. The Refusal Agent is enabled for the OSWorld-G benchmark, which includes a mix of feasible and infeasible tasks. For the ScreenSpot-Pro benchmark, the Refusal Agent is disabled, as all tasks in that benchmark are guaranteed to have valid targets.

A critical challenge in designing the Refusal Agent lies in balancing refusal accuracy against the False Positive Rate (FPR). As detailed in the Appendix, this presents a clear performance trade-off; for instance, one configuration can achieve a maximal refusal accuracy of 78.0\%, but at the cost of a 5.3\% FPR. Given this trade-off, we prioritized a reliable user experience by designing an agent that minimizes FPR. Our approach is based on the core insight that to overcome a VLM's inherent bias toward action, one must decouple the process of deliberation from the final decision. Our agent operationalizes this ``think before you act'' strategy using a multi-part, chain-of-thought prompt that compels the VLM to articulate its reasoning before making a judgment. This principled design achieves our desired balance: a strong 68.5\% refusal accuracy while maintaining a minimal FPR of just 3.3\%. As shown in Table~\ref{tab:refusal_summary}, this agentic, two-step reasoning process is significantly more effective than single-shot models, which conflate deliberation and action and thus struggle to reliably identify impossible tasks.

\begin{table}[tbp]
\centering
\caption{Refusal accuracy comparison on the OSWorld-G benchmark (N=54 infeasible tasks).}
\label{tab:refusal_summary}
\begin{tabular}{lc}
\toprule
\textbf{Model / Method} & \textbf{Refusal Acc. (\%)} \\
\midrule
Models (JEDI, UI-TARS, etc.) & 0.0 -- 18.5 \\
Gemini-2.5-Pro & 38.9 \\
\midrule
\textbf{Our Method (Refuser Agent)} & \textbf{68.5} \\
\bottomrule
\end{tabular}
\end{table}

\begin{figure}[tbp]
\centering
\includegraphics[width=\textwidth]{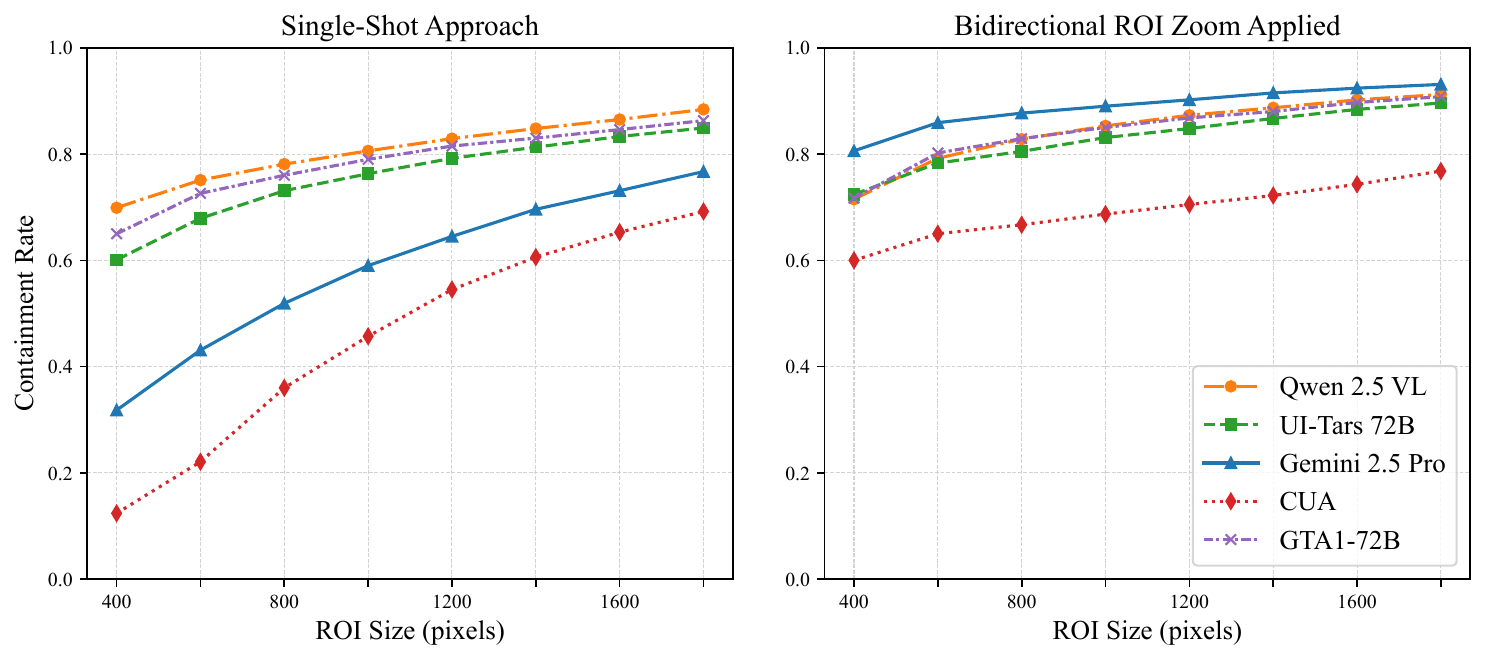}
\caption{Comparison of single-shot method performance (left) against that of our bidirectional ROI zooming method (right) on the ScreenSpot-Pro benchmark using various VLMs. Our method provides a substantial improvement in ROI-containment rate across all evaluated VLMs. The gains are most pronounced at smaller ROI sizes, where the adaptive zoom enables recovery from initial localization errors.}
\label{fig:ssp_adaptive_vs_static}
\end{figure}

\paragraph{ROI Deduction}

The most straightforward method for extracting an effective Region of Interest (ROI) from a full-screen image, given a grounding model, is to crop a fixed-size rectangle centered at the model's predicted grounding point. The result of this one-shot approach to ROI extraction is illustrated on the left side of Figure~\ref{fig:ssp_adaptive_vs_static}. As the ROI size increases, the likelihood of containing the target GUI element also increases, which is reflected in the positively sloped graphs. Interestingly, different models respond differently to varying ROI sizes. Notably, commercial models such as Gemini 2.5 Pro and CUA perform worse in terms of containment rate compared to baseline models and exhibit greater sensitivity to ROI size changes. This suggests that these models are optimized for specific screen resolutions and more narrowly defined tasks.

All VLMs show significant improvement in containment rate when using MEGA-GUI's bidirectional ROI Zoom method, as shown on the right side of Figure~\ref{fig:ssp_adaptive_vs_static}. To generate this result, the full-sized screen image is repeatedly reduced (and occasionally expanded, according to Algorithm~1) until the target ROI size is achieved. A key consideration in our iterative framework is the trade-off between search robustness and in-ference latency. While our research prototype averages 21.54 seconds (0.359 minutes) per task, this latency is an artifact of our experimental setup---which prioritizes methodological clarity over engineering optimization---and not a fundamental limitation of our algorithm. The overhead stems from sequential client-server API calls in a non-optimized environment. A production-level implementation, co-locating the agents and processing requests in batches, would dramatically reduce this latency and make the approach viable for interactive applications.

On the ScreenSpot-Pro benchmark, Gemini 2.5 Pro achieves a containment rate that scales from 72.6\% at a 400-pixel ROI to 93.2\% at 1800 pixels. Its performance is consistently superior to that of our method implemented with other VLMs such as Qwen 2.5 VL and UI-Tars 72B. Consequently, we select Gemini 2.5 Pro as the Stage 1 agent for our experiments, providing the basis for downstream processing in Stage 2. Similar trends are observed on the OSWorld-G benchmark datasets, with detailed results provided in the Appendix.

\paragraph{Fine-Grained Grounding}
\label{sec:stage2_analysis}

\begin{figure}[h]
\centering
\includegraphics[width=0.5\textwidth]{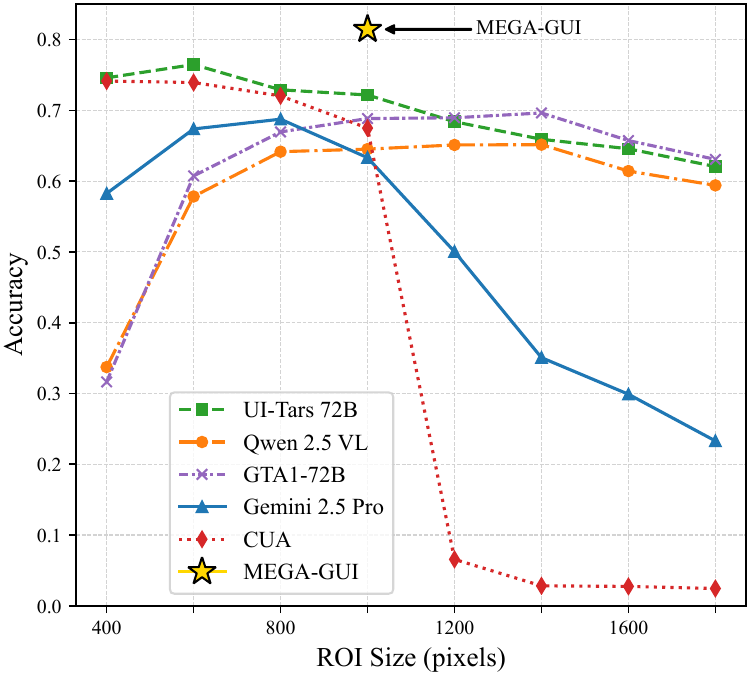}
\caption{Conditional Grounding Accuracy of various VLMs on the ScreenSpot-Pro benchmark. Accuracy is calculated only on ROIs that successfully contain the target element, isolating Stage 2 performance. The overall end-to-end accuracy is represented by the composite score (see Appendix).}
\label{fig:accuracy_score}
\end{figure}

Given the ROIs of a designated size produced by the Stage 1 method, we evaluated various VLMs as grounding agents to assess how accurately they can identify the target GUI element described in the user instruction within the ROI images. The experimental results are shown in Figure~\ref{fig:accuracy_score}, where accuracy is calculated using only the ``good'' ROIs---that is, those that contain the target element. A general downward slope is observed, indicating that smaller images make it easier to locate the target without being distracted by surrounding elements. Sudden drops in performance at smaller ROI sizes for some VLMs suggest a significant domain shift in the grounding task compared to their training data, which typically consists of full-screen or near-full-screen images rather than cropped partial views.

The choice of ROI size in MEGA-GUI must therefore balance two competing factors: the containment rate (i.e., whether the target is within the ROI) and the localization accuracy (i.e., how precisely the model identifies the target within the ROI), which generally increase and decrease, respectively, with larger ROI sizes. We define the product of these two metrics as the \emph{composite score}, plotted in the Appendix. This score tends to plateau around ROI sizes of 1000 pixels across most VLMs. Based on this observation, we adopt a fixed ROI size of 1000 pixels for the final grounding experiments on both the SSP and OSG benchmarks presented in this paper.

As shown in Figure~\ref{fig:accuracy_score}, UI-TARS-72B demonstrates the strongest baseline conditional accuracy at an ROI size of 1000 pixels. For optimization of Scale Agent, we further evaluated this model under different image scaling factors, ranging from 1$\times$ to 4$\times$, and found that a 3$\times$ scale yields the best improvement in grounding accuracy. We also tested various configurations of the instruction-rewriting agent. Ablation studies revealed that several strategies are effective, including enforcing structured JSON output formatting and providing the VLM with additional context---such as which application window is relevant to the user's instruction or visual cues indicating the target element. The combined effect of the Scale Agent and Rewrite Agent in our MEGA-GUI framework leads to a further improvement in localization accuracy compared to direct, unmodified calls to the base model. This results in a final conditional accuracy of 81.4\% on the ScreenSpot-Pro benchmark---indicated by the star-shaped marker in Figure~\ref{fig:accuracy_score}---demonstrating the high precision of our framework once the correct region is identified.

\section{DISCUSSION}

A central finding of our analysis is a ``No Free Lunch'' principle in VLM specialization. Models exhibit distinct and often inverse performance profiles across ROI scales: generalist models excel at broad contextual search, while specialist models perform best in fine-grained grounding. Some are better suited for ROI selection, others for precise localization. This insight---enabled by GBT---highlights the value of modularity and the need for adaptive, data-driven model routing.

Beyond these empirical findings, we also release the Grounding Benchmark Toolkit (GBT) to support systematic and reproducible research on GUI grounding. GBT enables resolution-based sweeps—for example, varying ROI sizes from 400 to 1800\,px—making it straightforward to compare different grounding models and agent configurations under a unified protocol. The toolkit additionally provides the test-set ROIs used in our experiments, extracted from the ScreenSpot-Pro and OSWorld-G benchmarks, allowing researchers to directly reproduce our results and interchange subcomponents within the MEGA-GUI framework.

Despite these advances, MEGA-GUI has certain limitations. Its reliance on multiple VLM queries introduces latency, although our efficiency analyses show this is a tunable trade-off through model selection and pruning strategies. The current framework also assumes access to high-performing proprietary models, which may limit deployment in resource-constrained settings. Future work should explore open-source alternatives or investigate model distillation techniques to produce smaller, highly efficient specialists. Additionally, while the Refuser Agent enhances safety by blocking infeasible or ambiguous commands, integrating more advanced ethical reasoning—such as detecting manipulative or harmful instructions—remains an important open challenge.

The broader impacts of this work include the potential to enable more capable assistive technologies for users with disabilities and more efficient RPA systems for enterprise automation. However, these same advancements raise important concerns about dual-use, such as applications in surveillance or the automation of malicious tasks. By open-sourcing our toolkit and promoting transparent evaluation, we aim to foster responsible development and empower the community to study, adapt, and mitigate these risks.

\subsection{Qualitative Failure Analysis}

While MEGA-GUI sets a new state of the art in quantitative performance, a qualitative inspection of its failure cases is essential for understanding the remaining challenges and guiding future research.

\paragraph{Failure of Search: Attentional Fixation.}
The first archetype involves failures of the Bidirectional ROI Zoom agent. On visually dense UIs with many repetitive, similar-looking elements (e.g., a layers panel in design software), the Stage~1 VLM can develop a fixation on a salient but incorrect region. This produces a non-productive loop in which the agent repeatedly returns the same wrong coordinates with high confidence. Although the bidirectional policy is designed to recover by zooming out, a sequence of high-confidence errors can deplete the error budget ($E_{max}$) before correction occurs, leading to premature termination.

\paragraph{Failure of Grounding: Deterministic Bias.}
The second archetype highlights limitations of the Stage~2 grounding VLM (UI-TARS-72B in our case). These errors reflect biases inherited from the model’s training history. For example, if the model has been frequently exposed to UIs where a three-dots menu represents ``settings,'' it will repeatedly and confidently select that icon---even when the user’s instruction and the current application clearly indicate a gear icon. Such path-dependent priors lead to predictable blind spots that our framework can mitigate but not completely eliminate.

\paragraph{Failure by Instructional Over-Correction.}
The third archetype occurs when the Context-Aware Rewrite Agent unintentionally makes an instruction more difficult for the grounding model to follow. For example, the simple command ``Save the document'' may enable the grounding model to identify a floppy-disk icon through strong visual priors. If rewritten as ``Click the button labeled `Save' next to the file name,'' however, the model may over-focus on the literal text ``Save,'' fail to find it, and return an error. This \textit{curse of specificity} highlights that the interface between agents is delicate: the best instruction is not always the most explicit, but rather the one aligned with the biases and expectations of the executing model.

\section{CONCLUSION}

In this paper, we introduced MEGA-GUI, a modular multi-agent framework that establishes a new state of the art in GUI grounding. Critically, our approach requires no model training or fine-tuning; instead, it orchestrates publicly available VLMs as specialized agents. By decomposing the grounding task into distinct subtasks—including robust visual search via a bidirectional ROI zoom algorithm and ambiguity resolution through a context-aware instruction rewriting agent—our framework demonstrates the synergistic benefit of composing models that are optimally suited for specialized roles. This principled, system-level architecture achieves 73.18\% accuracy on ScreenSpot-Pro and 68.63\% on OSWorld-G, validating its effectiveness. MEGA-GUI represents a meaningful shift toward more capable and principled autonomous agents, and to accelerate future research, we release our Grounding Benchmark Toolkit (GBT) to support systematic and transparent evaluation in this domain.

\bibliographystyle{apalike}
\bibliography{aaai2026_SJ} 

\clearpage
\onecolumn 
\appendix

\section{Guide to the Appendix}
\label{sec:guide}
This technical appendix provides the formal underpinnings and empirical validation for the MEGA-GUI framework. It begins with a rigorous exposition of the core Bidirectional ROI Zoom algorithm, complete with unabridged pseudocode, designed for iterative and precise element localization. To ensure full reproducibility, we then detail the experimental methodology, including the distinct challenges posed by the ScreenSpot-Pro and OSWorld-G benchmarks, our computing infrastructure, and specific hyperparameter configurations. The core of our empirical analysis resides in a granular breakdown of performance, quantifying the critical trade-off between Stage 1's ROI containment rate and Stage 2's grounding accuracy to establish an optimal operating point. Systematic ablation studies further deconstruct the framework's architecture, isolating the contributions of prompt engineering, image scaling, and search efficiency. The investigation extends beyond quantitative metrics to an advanced analysis of agent capabilities, illuminating both the model's capacity for instruction refusal and the qualitative failure archetypes that define current limitations. The appendix culminates in a forward-looking discussion on the broader impacts of this technology, from enhancing digital accessibility to shaping responsible AI paradigms.

\section{Algorithm Specifications and Formalisms}
\label{sec:appendix_framework}

This appendix details the technical specifications of the MEGA-GUI framework, with a primary focus on the Bidirectional ROI Zoom algorithm. We formalize the algorithm with unabridged pseudocode and provide precise definitions for all supporting subroutines.

\subsection{Overview of the Bidirectional ROI Zoom Algorithm}
The Bidirectional ROI Zoom algorithm is a closed-loop control mechanism designed to iteratively refine a Region of Interest (ROI) for precise element localization within a graphical user interface (GUI) screenshot. It improves upon traditional unidirectional zooming by incorporating bidirectional adjustments: an aggressive inward crop (zoom-in) for refinement and a controlled outward expansion (zoom-out) for error recovery.

The algorithm takes as input a full-screen image $I$ of dimensions $W \times H$, a natural language instruction $q$, and a Vision-Language Model (VLM) $\mathcal{M}$ that predicts a target coordinate $p=(x,y)$.

A key feature of the algorithm is its self-correcting mechanism, managed by an error counter $E$. This counter tracks the number of out-of-bounds predictions and triggers a corrective, forced zoom-in when a predefined threshold $E_{max}$ is reached. This prevents the agent from entering prolonged, unproductive search cycles.

The algorithm's behavior is governed by the following key hyperparameters:
\begin{itemize}
    \item $\Delta_{in} \in (0, 1)$: The reduction factor for the zoom-in operation.
    \item $\Delta_{out} \in (0, 1)$: The expansion factor for the zoom-out operation.
    \item $E_{max} \in \mathbb{N}^+$: The error count threshold that triggers a forced zoom-in.
    \item $S_{min} > 0$: The minimum ROI size (area proxy) that serves as a termination condition.
    \item $N_{stable} \in \mathbb{N}^+$: The number of recent predictions to consider for convergence.
    \item $\epsilon_{stable} > 0$: The maximum distance threshold for a cluster of predictions to be deemed stable, indicating convergence.
\end{itemize}

\subsection{Full Pseudocode}
The complete logic of the Bidirectional ROI Zoom algorithm is presented in Algorithm~\ref{alg:full_bidirectional_roi_zoom_cumulative}.

\begin{algorithm}[h!]
\caption{Bidirectional ROI Zoom}
\label{alg:full_bidirectional_roi_zoom_cumulative}
\begin{algorithmic}[1]
\REQUIRE Instruction $q$, Image $I$ with dimensions $W \times H$, Model $\mathcal{M}$
\REQUIRE Hyperparameters: $\Delta_{in}, \Delta_{out}, E_{max}, S_{min}, N_{stable}, \epsilon_{stable}$
\ENSURE Final cropped Region of Interest (bounding box)

\STATE $\text{ROI} \leftarrow [0, 0, W, H]$
\STATE $H \leftarrow \emptyset$ \COMMENT{History of in-bounds predictions}
\STATE $E \leftarrow 0$ \COMMENT{Out-of-bounds error counter}

\WHILE{$\text{GetArea}(\text{ROI}) > S_{min}$ \textbf{and} \textbf{not} $\text{SearchConverged}(H, N_{stable}, \epsilon_{stable})$}
    \STATE $\text{cropped\_image} \leftarrow \text{CropImage}(I, \text{ROI})$
    \STATE $p \leftarrow \mathcal{M}.\text{predict}(q, \text{cropped\_image})$
    
    \IF{$\text{IsInside}(p, \text{ROI})$}
        \STATE $H \leftarrow H \cup \{p\}$
        \STATE $\text{ROI} \leftarrow \text{ZoomIn}(\text{ROI}, p, \Delta_{in})$
    \ELSE
        \STATE $E \leftarrow E + 1$
        \IF{$E \geq E_{max}$}
            \STATE $\text{ROI} \leftarrow \text{ForcedZoomIn}(\text{ROI}, \Delta_{in})$
        \ELSE
            \STATE $\text{ROI} \leftarrow \text{ZoomOut}(\text{ROI}, \Delta_{out})$
        \ENDIF
    \ENDIF
\ENDWHILE

\STATE \textbf{return} $\text{Finalize}(\text{ROI}, H)$
\end{algorithmic}
\end{algorithm}

\subsubsection{Helper Functions.}
The subroutines invoked in Algorithm~\ref{alg:full_bidirectional_roi_zoom_cumulative} are defined as follows.

\begin{description}
    \item[\texttt{GetArea(ROI)}] Computes a proxy for the ROI's size by taking the maximum of its width and height. $\text{GetArea}([x_{min}, y_{min}, x_{max}, y_{max}]) = \max(x_{max} - x_{min}, y_{max} - y_{min})$.

    \item[\texttt{SearchConverged(H, N, $\epsilon$)}] Assesses convergence based on a sliding window of the last N predictions. It returns true if the most recent prediction is within the distance threshold $\epsilon$ of all other $N - 1$ predictions in the window, indicating that the search has stabilized on a single location.

    \item[\texttt{ZoomIn(ROI, p, $\Delta$)}] Performs an asymmetric zoom by moving the ROI boundaries inward toward the predicted point $p$. This focuses the search space. For example, if $p$ lies in the right half of the ROI, the left boundary is shifted inward.

    \item[\texttt{ForcedZoomIn(ROI, $\Delta$)}] Performs a symmetric zoom-in, reducing the ROI's size from all four sides. This conservative reduction is triggered when the out-of-bounds error count reaches the threshold $E_{max}$.

    \item[\texttt{ZoomOut(ROI, $\Delta$)}] Uniformly expands the ROI boundaries outward by a factor of $\Delta_{out}$. This operation is used to regain context following an out-of-bounds prediction.

    \item[\texttt{IsInside(p, ROI)}] Returns true if point $p=(x,y)$ is within the ROI's bounds, i.e., $x_{min} \leq x \leq x_{max}$ and $y_{min} \leq y \leq y_{max}$.

    \item[\texttt{Finalize(ROI, H)}] Computes the final bounding box. If the search converged (i.e., \texttt{SearchConverged} is true), this function may return an ROI centered on the mean of the stable points in $H$. Otherwise, it returns the last computed ROI.

    \item[\texttt{EuclideanDist($p_1, p_2$)}] Calculates the standard Euclidean distance: $\sqrt{(x_1 - x_2)^2 + (y_1 - y_2)^2}$.
\end{description}

\section{Experimental Setup for Reproducibility}
\label{sec:setup}

\subsection{Benchmark Dataset Descriptions}
We evaluate our framework on two distinct benchmarks to ensure a comprehensive assessment of its capabilities. 
\begin{itemize}
    \item \textbf{ScreenSpot-Pro (SSP)} : This benchmark comprises 1,581 tasks on high-resolution (4K) interfaces of professional software. It is specifically designed to test an agent's spatial precision and robustness to extreme visual density, where target elements are often minuscule and surrounded by distractors.
    \item \textbf{OSWorld-G (OSG)} : This benchmark includes 564 tasks on standard 1080p operating system interfaces. It focuses on evaluating an agent's functional "software commonsense" and its ability to handle semantic ambiguity in everyday GUI environments.
\end{itemize}
Our primary evaluation metric is Top-1 Accuracy, where a prediction is correct if its coordinate falls within the ground-truth element's annotated region (a bounding box for SSP and a bounding polygon for OSG).

\subsection{Computing Infrastructure and Software}
All experiments were conducted using \textbf{Python 3.11} on a server equipped with eight \textbf{NVIDIA V100 GPUs}. The multi-agent framework was orchestrated using key libraries from the LangChain ecosystem, including \textbf{LangGraph (v0.3.24)} and \textbf{LangChain (v0.3.22)}.

We utilized two primary proprietary foundation models accessed via their official APIs: \textbf{GPT-4o} (model version \texttt{gpt-4o-1120}) via the OpenAI Python client (v1.70.0), and the latest available version of \textbf{Gemini 2.5 Pro}. The core software stack included libraries such as NumPy (v1.26.4), Pandas (v2.2.3), and Pillow (v11.0.0). A complete list of packages and their versions is provided in our public code repository to ensure full reproducibility.

\subsection{Hyperparameter Settings}
The key hyperparameters used across all experiments for the MEGA-GUI framework are detailed in Table~\ref{tab:hyperparameters}. These values were held constant to ensure fair and consistent comparisons.

\begin{table}[h!]
\centering
\caption{Key hyperparameters for the Bidirectional ROI Zoom algorithm.}
\label{tab:hyperparameters}
\begin{tabular}{ll}
\toprule
\textbf{Hyperparameter} & \textbf{Value} \\
\midrule
$\Delta_{in}$ (Zoom-in Factor) & 10 \% \\
$\Delta_{out}$ (Zoom-out Factor) & 5 \% \\
$E_{max}$ (Error Threshold) & 5 \\
$S_{min}$ (Min. Area Threshold) & $1000 \times 1000$ pixels \\
$N_{stable}$ (Convergence Window) & 3 \\
$\epsilon_{stable}$ (Convergence Radius) & 50 pixels \\
\bottomrule
\end{tabular}
\end{table}

\subsection{Code and Data Availability}
To uphold our commitment to reproducible research and foster further innovation, all code, experimental data, and configuration files necessary to replicate our findings are publicly available in our open-source repository: \url{https://github.com/samsungsds-research-papers/mega-gui}.

The release includes commented Jupyter notebooks that implement the full MEGA-GUI pipeline for both the ScreenSpot-Pro and OSWorld-G benchmarks. It also contains the complete source code for our \textbf{Grounding Benchmark Toolkit (GBT)}, including data pre-processing scripts and evaluation harnesses. To accelerate follow-up research, we provide the complete set of pre-computed Region-of-Interest (ROI) datasets generated by our Bidirectional ROI Zoom experiments across all tested VLMs. This allows researchers to build upon our Stage 1 results without incurring redundant computational costs. Finally, detailed guides for environment setup, API key configuration, and deploying open-source models with our vLLM-based serving solution are provided to ensure a straightforward replication process.

\section{Detailed Experimental Results}
\label{app:detailed_results}
This appendix provides a comprehensive breakdown of our experimental results, expanding on the findings presented in the main paper. We present full performance metrics across all tested grounding agents, followed by in-depth analyses of ROI deduction, fine-grained grounding, and composite scores on both the ScreenSpot-Pro (SSP) and OSWorld-G (OSG) benchmarks.
\subsection{Main Performance Results}
\label{ssec:main_results}
Table~\ref{tab:main_results_supp} provides a consolidated view of grounding accuracy on the two standard benchmarks---ScreenSpot-Pro (SSP) and OSWorld-G (OSG)---when different Vision-Language Models (VLMs) act as the \emph{Stage 2 grounding agent} within our MEGA-GUI framework. All runs share an identical \emph{Stage 1} (Gemini 2.5 Pro with our bidirectional zoom to a $1000 \times 1000$ ROI) and identical post-processing (3$\times$ bicubic scaling and context-aware rewrite). The table confirms the advantage of the decoupled design: UI-TARS-72B sets a new state of the art on both datasets, validating our choice of this model as the default Stage 2 specialist.

\begin{table}[h]
\centering
\caption{Grounding accuracy (\%) on SSP and OSG under MEGA-GUI's optimal configuration. All figures are Top-1.}
\label{tab:main_results_supp}
\sisetup{table-align-text-post=false}
\begin{tabular}{l S[table-format=2.2] S[table-format=2.2]}
\toprule
\textbf{Stage 2 Grounding Agent} & {\textbf{SSP}} & {\textbf{OSG}} \\
\midrule
CUA (Operator) & 37.76 & 21.63 \\
Gemini 2.5 Pro & 46.86 & 42.73 \\
UI-TARS-7B & 69.82 & 63.29 \\
GTA1-7B & 68.75 & 64.89 \\
Qwen-VL-2.5-72B & 71.23 & 66.84 \\
GTA1-72B & 72.35 & 68.08 \\
\textbf{UI-TARS-72B} & \textbf{73.18} & \textbf{68.63} \\
\bottomrule
\end{tabular}
\end{table}

These results reinforce our core claim: decoupling coarse spatial localization from fine-grained grounding enables each agent to exploit its comparative advantage, outperforming any monolithic alternative.
\subsubsection{Stage 1 ROI Deduction: Containment Analysis}
\label{ssec:roi_containment}
\begin{figure}[t]
\centering
\includegraphics[width=1\textwidth]{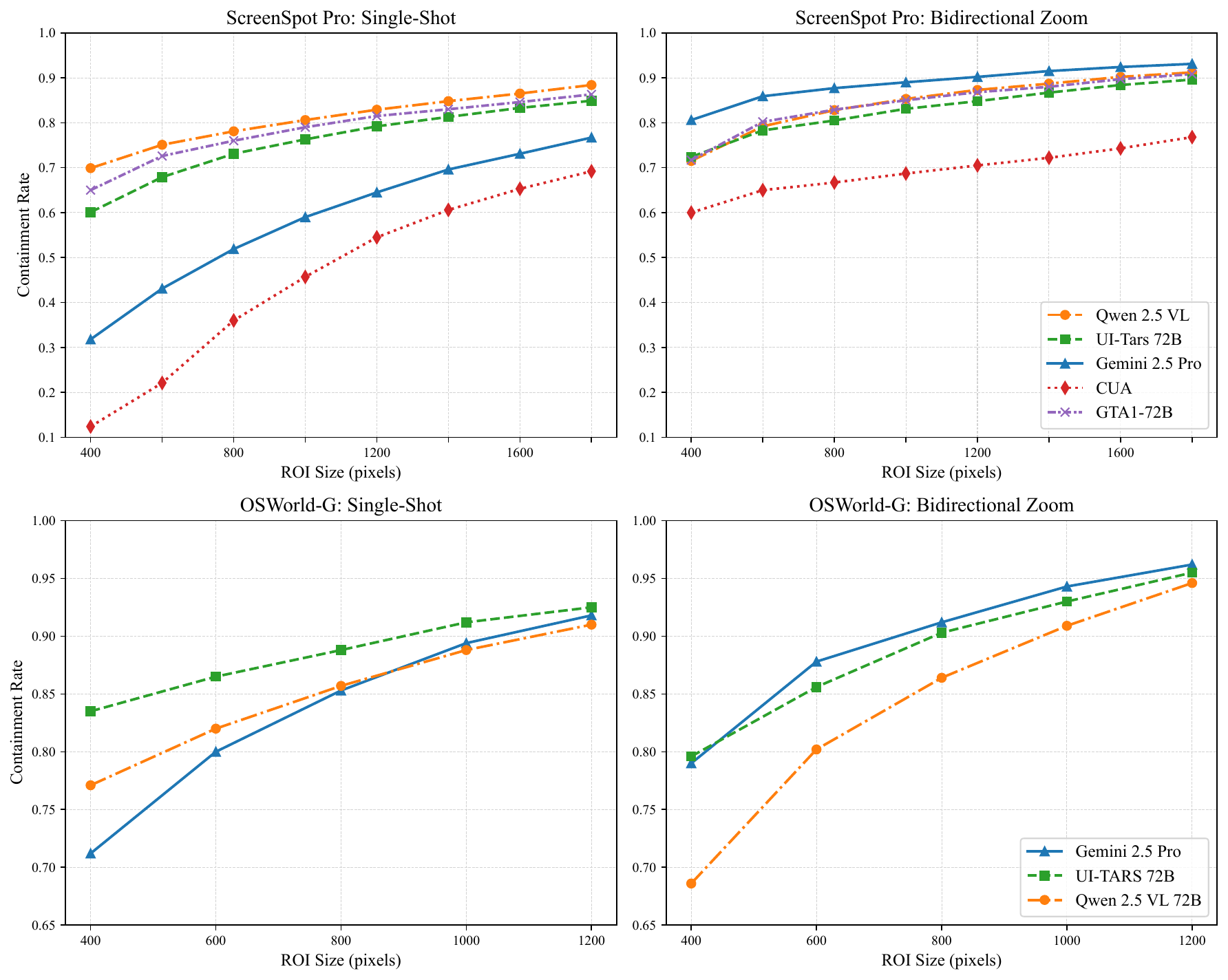}
\caption{Containment--ROI curves for SSP (top) and OSG (bottom), contrasting static one-shot cropping (left) with bidirectional zoom (right).}
\label{fig:containment_curves}
\end{figure}

We quantify Stage 1 performance using the \emph{containment rate}---the fraction of test cases in which the Region-of-Interest (ROI) fully encloses the ground-truth element---swept over a range of square ROI sizes.\footnote{SSP: 400--1800,px in 200-px steps; OSG: 400--1200,px in 200-px steps.} Figure~\ref{fig:containment_curves} compares a static one-shot baseline (ROI centered on the model's first prediction) with our bidirectional zoom driven by Gemini 2.5 Pro.
\paragraph{ScreenSpot-Pro} Gemini attains an average containment of \textbf{88.8\%}---rising from 80.6\% at a 400-px crop to 93.1\% at 1800-px---outperforming the best specialist (Qwen-VL-2.5-72B) by $\sim$7 percentage points (pp) in the low-context regime. Bidirectional zoom delivers a dramatic uplift of up to \textbf{+48.8 pp} at 400 px, effectively recovering from early mis-centered guesses and enabling precise localization at the small ROIs required for Stage 2 grounding.
\paragraph{OSWorld-G} Containment starts higher overall: 79.0\% at 400 px and peaks at 96.2\% for 1200 px, averaging \textbf{89.7\%}. Because OSG screens are less cluttered but instructions are more ambiguous, zoom gains are modest and static crops occasionally fare better at the smallest sizes---an effect we mitigate with adaptive zoom activation.
The dataset-specific patterns underscore the benefit of MEGA-GUI's modular design: high-resolution, icon-dense interfaces profit from aggressive zooming, whereas mid-resolution semantic tasks require a softer strategy. These insights directly inform the stage-wise model routing policy.

\subsection{Balancing Containment and Accuracy via Composite Score}
\label{sec:stage2_analysis_merged}

\begin{figure}[h!]
    \centering
    \includegraphics[width=0.5\textwidth]{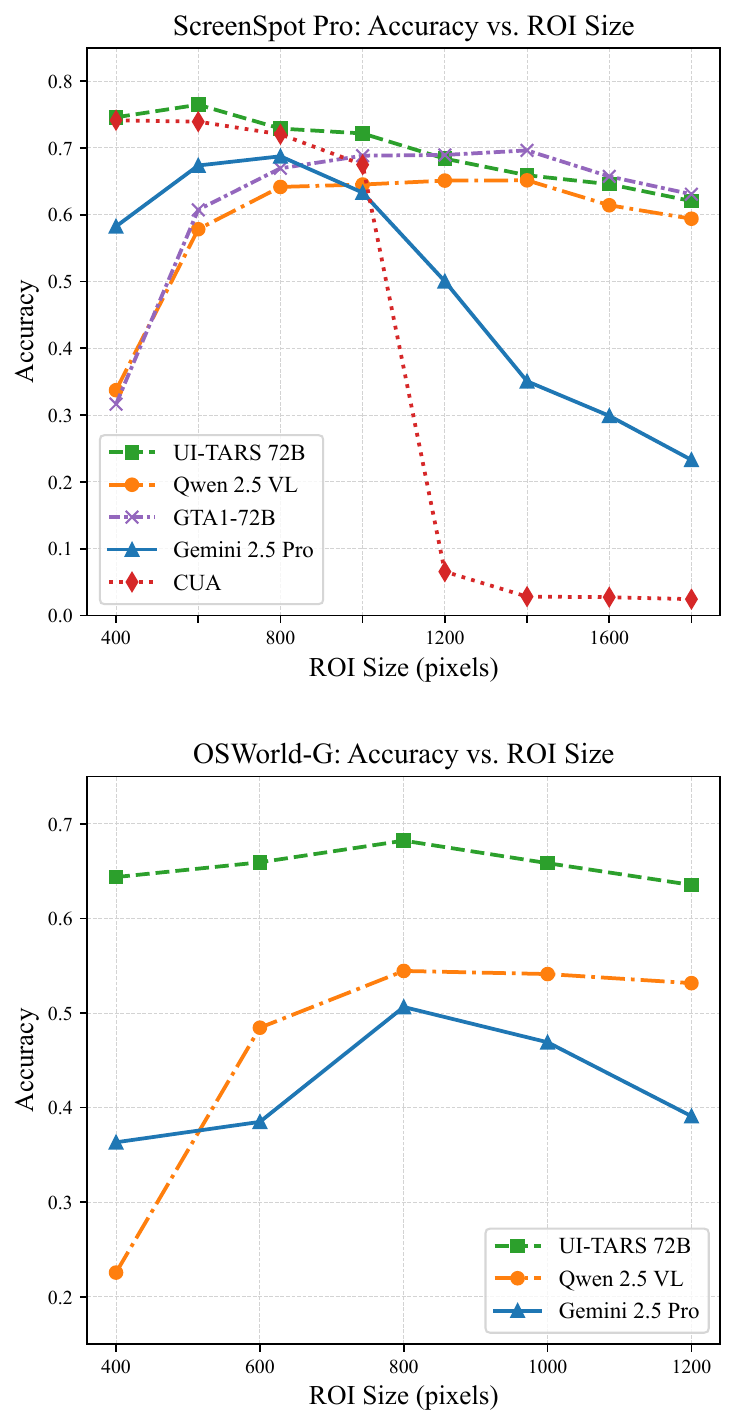}
    \caption{Grounding accuracy vs.\ ROI size for SSP (top) and OSG (bottom). Accuracy generally decreases as ROIs get larger and more cluttered, creating a trade-off with Stage 1's containment rate.}
    \label{fig:accuracy_score_supp}
\end{figure}

Given the Regions of Interest (ROIs) produced by Stage 1, we first evaluate Stage 2 performance by measuring the \emph{grounding accuracy} of various VLMs. This metric, calculated only on ROIs that correctly contain the target element, is plotted against ROI size in Figure~\ref{fig:accuracy_score_supp}. For most models, a clear downward trend is visible: smaller, less cluttered ROIs facilitate more precise localization. This initial analysis, however, reveals a critical trade-off: Stage 1's containment rate improves with larger ROIs, while Stage 2's grounding accuracy generally improves with smaller ones.

\begin{figure}[h!]
    \centering
    \includegraphics[width=0.5\textwidth]{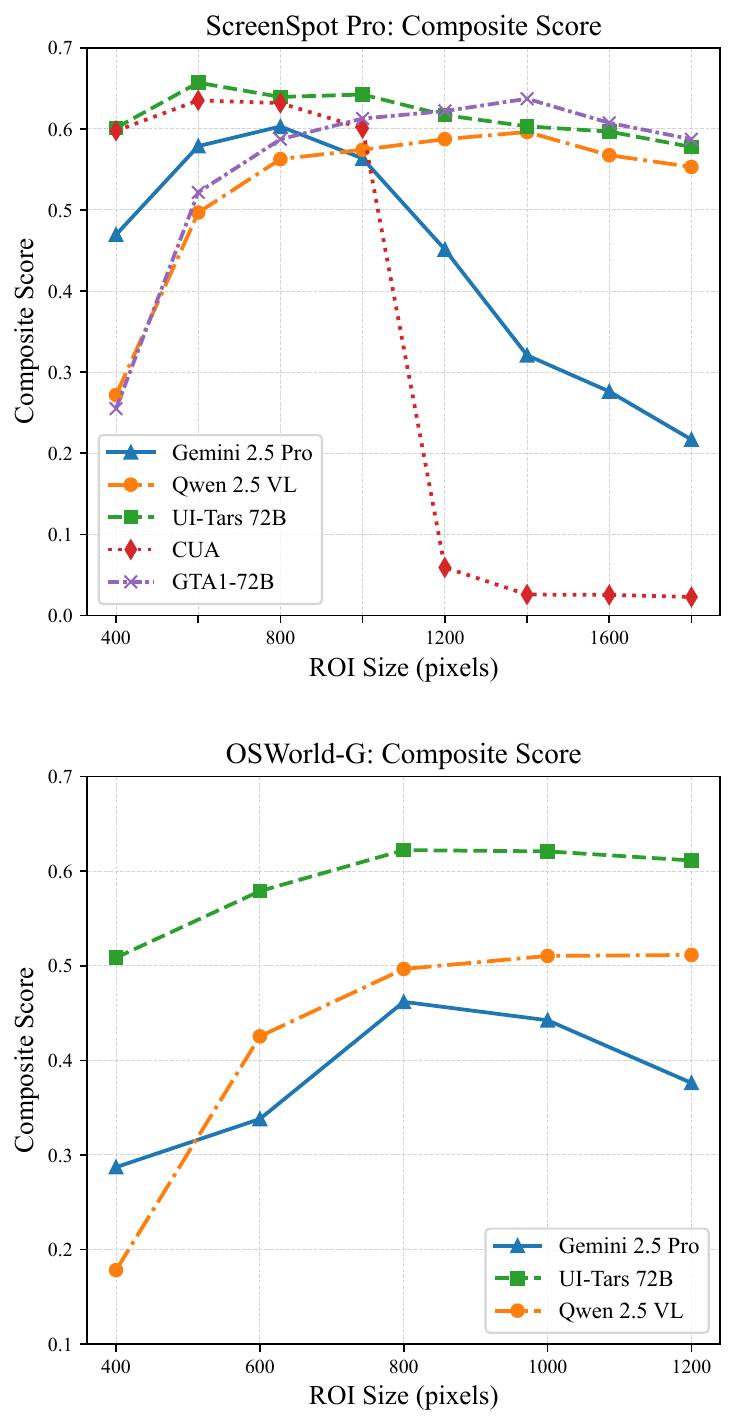}
    \caption{Composite score ($\text{containment rate} \times \text{grounding accuracy}$) vs.\ ROI size for SSP (top) and OSG (bottom). The score plateaus near 1000 px for top models, justifying it as our default target size.}
    \label{fig:composite_score_supp}
\end{figure}

To balance these competing factors, we define a \textbf{composite score}, calculated as the product of the Stage 1 containment rate and the Stage 2 grounding accuracy (i.e., $\text{containment rate} \times \text{grounding accuracy}$). This score, shown in Figure~\ref{fig:composite_score_supp}, represents the overall end-to-end probability of successfully grounding an element and allows us to identify an optimal operating point. The analysis reveals a clear "No Free Lunch" principle, where different VLMs exhibit distinct performance profiles depending on the dataset and ROI size.

\paragraph{Analysis on SSP (High-Res UIs).} On ScreenSpot-Pro (Figure~\ref{fig:composite_score_supp}, top), most models reach their peak composite score between 600 and 1000 pixels. \textbf{UI-TARS-72B} proves to be \emph{scale-robust}, maintaining a high score ($\ge 0.60$) across this entire band. In contrast, other models are more specialized. \textbf{CUA} excels as a "small-crop specialist," competitive up to ~800 px before its performance collapses, indicating sensitivity to visual dilution. \textbf{Gemini 2.5 Pro} shows an inverted-U curve, peaking at 800 px, while the "context-hungry" \textbf{Qwen-VL-2.5-72B} improves monotonically to a peak at a larger 1200-pixel ROI.

\paragraph{Analysis on OSG (Semantic UIs).} On OSWorld-G (Figure~\ref{fig:composite_score_supp}, bottom), the optimal ROI sizes shift rightward to the 900--1200 pixel range. Here, additional context helps disambiguate functional instructions. \textbf{UI-TARS-72B} again demonstrates its scale robustness by maintaining a high plateau. \textbf{Qwen-VL-2.5-72B} continues to benefit from more context, peaking near the largest ROI sizes. \textbf{Gemini 2.5 Pro} also peaks at 800 px but decays more mildly, showing less penalty for excess context in this lower-density setting.

\paragraph{Design Implications.} The composite score analysis provides a strong empirical basis for our system design. Across both datasets, the scores for top-performing models tend to plateau around a \textbf{1000-pixel ROI}. We therefore adopt this as the standard target size for MEGA-GUI, as it provides a robust balance between Stage 1 containment and Stage 2 accuracy. At this size, \textbf{UI-TARS-72B} consistently emerges as the strongest and most reliable grounding agent. As a final step, further optimizations from our specialized agents—the Conservative Scale Agent (3$\times$ upscaling) and the Context-Aware Rewrite Agent (instruction clarification)—significantly boost the performance of UI-TARS-72B, contributing to our state-of-the-art results.

\section{Ablation Studies and Component Contributions}
\label{sec:ablation}
This section is dedicated to a series of ablation studies designed to isolate and quantify the contribution of each key component within the MEGA-GUI framework. Each study targets a specific agent or parameter to understand its impact on overall performance and efficiency.

\subsection{Ablation Study on Prompt Engineering for Semantic Refinement}
\label{app:ablation_prompting}

This appendix provides the detailed experimental protocol, complete prompt formulations, and full quantitative results for the instruction rewriting ablation study referenced in the main text. The objective of this study was to systematically evaluate how different inference-time prompting strategies impact the grounding accuracy of a frozen Vision-Language Model (VLM), thereby validating the design of the Context-Aware Rewrite Agent.

\subsubsection{Experimental Protocol}
The experiment was conducted under the following controlled conditions to ensure reproducibility and isolate the impact of prompt engineering.

\begin{itemize}
    \item \textbf{Rewriting Model:} GPT-4o (model version \texttt{gpt-4o-1120}) was used to perform the instruction rewriting task.
    \item \textbf{Grounding Model:} UI-TARS-72B with frozen weights served as the downstream grounding agent. No model training or fine-tuning was performed.
    \item \textbf{Dataset:} Our evaluation subset—a 100-instance stratified sample from ScreenSpot-Pro—is included in the GBT to ensure full reproducibility.
    \item \textbf{Evaluation Metric:} Top-1 Grounding Accuracy. A prediction was deemed correct if the predicted coordinate fell within the ground-truth bounding box of the target UI element.
    \item \textbf{Image Conditions:} Each prompt strategy was evaluated on the cropped ROI images under two distinct visual fidelity conditions:
    \begin{itemize}
        \item \textbf{Raw Resolution:} The original, unmodified image crops from the dataset.
        \item \textbf{High Resolution:} The same images upscaled by a factor of 2$\times$ using bicubic interpolation.
    \end{itemize}
\end{itemize}

\subsubsection{Prompt Formulations}
We evaluated five distinct strategies for rewriting the user's raw instruction. The baseline condition (`Raw Instruction`) uses the original instruction with no modifications. The f-string templates below were used with the rewriting model.

\begin{listing}[tb]
\caption{Context Injection Prompt Template}
\label{lst:prompt_two_stage}
\begin{lstlisting}[language=Python]
f"""You have a screenshot and an instruction.
Follow these two steps:
1. Identify the application (brief reasoning).
2. Within that app, locate the UI element matching:
   "{raw_instruction}"
Respond exactly as:
Application: <app_name>
Element: <description or selector>
"""
\end{lstlisting}
\end{listing}

\begin{listing}[tb]
\caption{Spatio-Visual Description Prompt}
\label{lst:prompt_vsd}
\begin{lstlisting}[language=Python]
f"""Identify the most relevant and precisely located UI element for the following instruction.
Follow these rules:
- Describe the element type (e.g. button, dropdown, icon).
- Mention its visual traits: shape, color, text label, border, icon.
- Explain its spatial relationships to other UI elements (e.g., to the left of X, below Y).
Instruction: {raw_instruction}"""
\end{lstlisting}
\end{listing}

\begin{listing}[tb]
\caption{Disambiguation Prompt}
\label{lst:prompt_eld}
\begin{lstlisting}[language=Python]
f"""You are given a screenshot and an instruction. Your task is to identify the most relevant UI element for a click action.
Be specific and unambiguous. Avoid vague descriptions like "the icon" or "the button".
Only visible elements in the screenshot should be considered.
Instruction: {raw_instruction}"""
\end{lstlisting}
\end{listing}

\begin{listing}[tb]
\caption{Structured Output (JSON) Prompt}
\label{lst:prompt_json}
\begin{lstlisting}[language=Python]
f"""Identify the most relevant UI element for the given instruction.
Return your answer in the following JSON format:
{{
    "clarified_instruction": "<rewritten instruction in one sentence>",
    "element_description": "<detailed description of the target UI element>"
}}
Instruction: {raw_instruction}"""
\end{lstlisting}
\end{listing}

The \textbf{Hybrid Strategy} combined the principles from all the above prompts into a single, comprehensive instruction for the VLM.

\subsubsection{Full Results and Discussion}
The performance gains for all strategies across both image conditions are presented in Table~\ref{tab:appendix_rewrite_ablation}.

\begin{table}[t]
\centering
\begin{threeparttable}
  \caption{Impact ($\Delta$, in pp) of instruction-rewriting strategies on grounding accuracy, tested on SSP with raw vs.\ 2$\times$ upscaled images. Hybrid yields the largest gains across both.}
  \label{tab:appendix_rewrite_ablation}
  \sisetup{table-format=+1.2}
  \begin{tabular}{l S[table-format=+1.2] S[table-format=+1.2]}
    \toprule
        \textbf{Strategy} & {\makecell{\textbf{$\Delta$ High} \\\textbf{(pp)}}} & {\makecell{\textbf{$\Delta$ Raw} \\\textbf{(pp)}}} \\
    \midrule
    Context Injection      & {+2.00} & {+1.00} \\
    Spatio-Visual          & {+1.00} & {+4.00} \\
    Disambiguation         & {+0.00} & {+3.00} \\
    Structured Output      & {+1.00} & {+6.00} \\
    \midrule
    \textbf{Hybrid}        & \bfseries {+2.00} & \bfseries {+6.00} \\
    \bottomrule
  \end{tabular}
  \begin{tablenotes}
    \item[*] \textit{Measured relative to baselines of 68.00\% (High) and 61.00\% (Raw).}
  \end{tablenotes}
\end{threeparttable}
\end{table}

The results confirm the analysis presented in the main text. Strategies that enforce structured reasoning (Structured Output and Hybrid) provide the most significant performance lift on raw-resolution images, where visual information is less distinct. The performance dichotomy of the Spatio-Visual prompt—beneficial for raw resolution but detrimental for high resolution—underscores the importance of tailoring prompt complexity to input fidelity. The Hybrid strategy's strong, consistent performance across both conditions validates its selection as the architectural basis for our final Context-Aware Rewrite Agent.

\subsection{Ablation: Impact of Image Scaling}

While the Bidirectional ROI Zoom agent effectively localizes the target region, the final fine-grained grounding step is highly sensitive to the visual fidelity of the input crop. To quantify this effect and optimize our pipeline, we conducted an ablation study on the impact of image scaling. This experiment evaluates how upscaling the Region of Interest (ROI) before passing it to the final grounding agent affects accuracy.

The study was performed on the complete ScreenSpot-Pro benchmark (1,581 tasks), using the 1000-pixel ROIs generated by our optimized Gemini 2.5 Pro-based zoom agent. We then passed these ROIs to our grounding agent, UI-TARS-72B, under various bicubic interpolation scaling factors (2$\times$, 3$\times$, and 4$\times$). The analysis was conducted under two parallel conditions: one using the original user instruction, and another using the enhanced instruction from our Context-Aware Rewrite Agent. The results, presented in Table~\ref{tab:scaling_ablation}, measure the performance gains relative to the native 1$\times$ resolution baseline.

\begin{table}[h!]
\centering
\begin{threeparttable}
    \caption{Gains in grounding accuracy (pp) from ROI up-scaling. The 3$\times$ factor provides the most consistent and significant improvement across both instruction conditions.}
    \label{tab:scaling_ablation}
    \sisetup{table-format=+2.2}
    \begin{tabular}{
      l
      S
      S
    }
    \toprule
    \textbf{Scale} & {\shortstack{\textbf{$\Delta$ Original Instr.} \\ \textbf{(pp)}}} & {\shortstack{\textbf{$\Delta$ Rewritten Instr.} \\ \textbf{(pp)}}} \\
    \midrule
    $2\times$ & {+2.22} & {+2.22} \\
    \textbf{3$\times$} & \bfseries {+2.40} & \bfseries {+2.34} \\
    $4\times$ & {+2.34} & {+2.53} \\
    \bottomrule
    \end{tabular}
    \begin{tablenotes}
      \item[*] \textit{Deltas are measured from the native 1$\times$ resolution baseline. Baseline accuracies (UI-TARS-72B grounding) are 65.59\% with the original instruction and 70.84\% with the rewritten instruction.}
    \end{tablenotes}
\end{threeparttable}
\end{table}

The results demonstrate that upscaling the ROI provides a clear and consistent performance benefit. A \textbf{3$\times$ scaling factor emerges as the optimal configuration}, delivering a robust accuracy improvement of \textbf{+2.40 pp} with original instructions and \textbf{+2.34 pp} with rewritten instructions.

This improvement can be attributed to the increased feature differentiability for the grounding VLM. Upscaling effectively "magnifies" the visual details of UI elements, making subtle cues like text characters, icon borders, and color gradients more distinct. This allows the model to better discriminate between the target element and nearby distractors, a common failure mode in dense interfaces. While 4$\times$ scaling shows a slight further gain with rewritten instructions, the 3$\times$ factor provides a more balanced and substantial improvement in both conditions, justifying its selection for our final MEGA-GUI pipeline as the Conservative Scale Agent's default setting.

\subsection{Ablation: Impact of Pruning on Search Efficiency}
\label{sec:ablation_pruning}

The overall efficiency and success of our framework are determined by the interplay between the chosen VLM agent, its algorithmic configuration, and the complexity of the UI. Our ablation studies are designed to demonstrate the robustness of our approach across these dimensions and to validate the strategic design choices of MEGA-GUI, with detailed results in Table~\ref{tab:10pruning_efficiency_appendix}, Table~\ref{tab:efficiency_pruned_appendix_restyled}, and Table~\ref{tab:convergence_resolution_clustered_avg_supp}.

\paragraph{A Spectrum of Specialized Capabilities.}
First, we establish baseline performance using the default $10\%$ pruning rate, as shown in Table~\ref{tab:10pruning_efficiency_appendix}. The results highlight the diverse spectrum of capabilities available to our modular framework. On the demanding ScreenSpot-Pro benchmark, Gemini-2.5-Pro sets a high bar for accuracy with an $89.7\%$ pass rate, demonstrating its strength in meticulous, high-fidelity search. In parallel, other models like Qwen-VL-2.5-72B and GTA1-7B offer highly efficient convergence pathways, successfully completing tasks in fewer than five steps on average. This diversity is a core strength, allowing for strategic agent selection based on task requirements—whether the priority is maximum precision or rapid execution.

\paragraph{Boosting Efficiency via Optimized Pruning.}
As stated in the main paper, a key hypothesis is that we can enhance efficiency by using a larger deduction rate with only a minor impact on performance. The results in Table~\ref{tab:efficiency_pruned_appendix_restyled} decisively validate this claim. By increasing the pruning rate from $10\%$ to $20\%$, we achieve substantial gains in efficiency across our top-performing models. For agents like Qwen-VL-2.5-72B and UI-TARS-72B, this optimization reduces the required steps while pass rates remain exceptionally stable, fluctuating by less than one percentage point. Even for our most accurate agent, Gemini-2.5-Pro, increasing the pruning rate delivers a significant improvement in convergence speed while its pass rate remains highly competitive. This confirms that the pruning rate is an effective lever for tuning performance, allowing us to optimize for speed without sacrificing the high accuracy that characterizes our framework.

\paragraph{Robustness Across Display Resolutions.}
Beyond algorithmic tuning, a successful agent must be robust to the wide range of screen resolutions found in real-world scenarios. We analyzed our framework's performance across various resolutions, with results clustered in Table~\ref{tab:convergence_resolution_clustered_avg_supp}. The findings show that our agent performs effectively across all tested environments, from standard HD to extreme 4K+ displays. As expected, task complexity scales with screen resolution, requiring more computational steps to ensure accurate convergence on larger canvases. This predictable scaling powerfully validates the core design of our bidirectional ROI zoom agent. The inherent difficulty of localizing a minuscule target on a high-resolution screen underscores the limitations of single-shot approaches. By intelligently decomposing a single, complex search into a series of more manageable steps, our framework excels precisely where others falter, maintaining high performance and demonstrating the superiority of a structured, multi-stage approach.

In summary, our analyses confirm that the MEGA-GUI framework is both powerful and flexible. It can leverage a diverse portfolio of specialized models, can be tuned for greater efficiency without significant compromise, and is architecturally designed to handle the challenges of modern, high-resolution user interfaces.

\begin{table}[h!]
\centering
\caption{Baseline Efficiency at 10\% Pruning.}
\label{tab:10pruning_efficiency_appendix}
\sisetup{table-format=2.2, round-mode=places, round-precision=2}
\begin{tabular}{ l S S[table-format=2.1] }
\toprule
\textbf{Model} & {\textbf{Avg. Steps}} & {\textbf{Pass Rate (\%)}} \\
\midrule
\multicolumn{3}{c}{\textit{ScreenSpot-Pro (High-Density, 1581 samples)}} \\
\midrule
GTA1-7B        & 4.62  & 81.1 \\
Qwen-VL-2.5-72B   & 4.71  & 85.0 \\
UI-TARS-72B & 6.47  & 82.5 \\
CUA           & 9.87  & 60.0 \\
Gemini-2.5-Pro    & 12.15 & 89.7 \\
\addlinespace
\midrule
\multicolumn{3}{c}{\textit{OSWorld-G (Functional, 510 samples)}} \\
\midrule
UI-TARS-72B & 3.43  & 90.2 \\
Qwen-VL-2.5-72B   & 3.45  & 88.2 \\
GTA1-72B    & 3.63  & 88.8 \\
GTA1-7B        & 3.60  & 84.5 \\
CUA           & 4.55  & 88.8 \\
Gemini-2.5-Pro    & 5.81  & 93.3 \\
\bottomrule
\end{tabular}
\end{table}

\begin{table}[h!]
\centering
\caption{Efficiency metrics by pruning level. ScreenSpot-Pro evaluated on 1,581 samples; OSWorld-G on 510.}
\label{tab:efficiency_pruned_appendix_restyled}
\setlength{\tabcolsep}{4pt}
\begin{tabular}{c c c}
\toprule
\multicolumn{3}{c}{\textbf{ScreenSpot-Pro (High-Density, 1,581 samples)}} \\
\midrule
\textbf{Pruning (\%)} & \textbf{Avg. Steps} & \textbf{Pass Rate (\%)} \\
\midrule
\multicolumn{3}{l}{\textbf{Gemini-2.5-Pro}} \\
10 & 12.15 & 89.7 \\
20 &  8.25 & 86.3 \\
30 &  6.61 & 81.3 \\
\addlinespace
\multicolumn{3}{l}{\textbf{Qwen-2.5-VL-72B}} \\
10 &  4.71 & 85.0 \\
20 &  4.27 & 84.3 \\
30 &  3.95 & 84.3 \\
\addlinespace
\multicolumn{3}{l}{\textbf{UI-TARS-72B}} \\
10 &  6.47 & 82.5 \\
20 &  5.29 & 82.8 \\
30 &  4.64 & 82.0 \\
\midrule
\multicolumn{3}{c}{\textbf{OSWorld-G (Functional, 510 samples)}} \\
\midrule
\multicolumn{3}{l}{\textbf{Qwen-2.5-VL-72B}} \\
10 &  3.45 & 88.2 \\
20 &  2.98 & 87.6 \\
30 &  3.78 & 91.2 \\
\addlinespace
\multicolumn{3}{l}{\textbf{UI-TARS-72B}} \\
10 &  3.43 & 90.2 \\
20 &  3.43 & 90.2 \\
30 &  3.00 & 90.0 \\
\addlinespace
\multicolumn{3}{l}{\textbf{GTA1-72B}} \\
10 &  3.63 & 88.8 \\
20 &  3.30 & 88.4 \\
30 &  2.93 & 88.4 \\
\bottomrule
\end{tabular}
\end{table}

\begin{table}[h!]
\centering
\caption{Detailed breakdown of convergence performance by resolution, grouped into logical clusters with weighted averages. The data is sorted by total pixel count to show performance trends.}
\label{tab:convergence_resolution_clustered_avg_supp}
\small 
\setlength{\tabcolsep}{4pt} 
\begin{tabular}{lcccc}
\toprule
\textbf{Resolution (W×H)} & \textbf{Total Pixels (M)} & \textbf{Avg. Steps} & \textbf{Pass Rate (\%)} & \textbf{Samples} \\
\midrule
\multicolumn{5}{l}{\textit{\textbf{Standard (HD/QHD)}}} \\
\midrule
(1920, 1080) & 2.07 & 8.05  & 100.00 & 19  \\
(2160, 1440) & 3.11 & 6.90  & 96.97  & 99  \\
(2560, 1440) & 3.69 & 8.62  & 93.97  & 514 \\
(2560, 1600) & 4.10 & 8.43  & 100.00 & 7   \\
(2560, 1664) & 4.26 & 8.77  & 95.83  & 48  \\
\cmidrule(lr){2-5}
\textbf{Average} & \textbf{3.61} & \textbf{8.36} & \textbf{94.75} & \\
\midrule
\multicolumn{5}{l}{\textit{\textbf{High-Res}}} \\
\midrule
(2880, 1800) & 5.18 & 11.10 & 93.90 & 82  \\
(2992, 1870) & 5.59 & 13.00 & 75.00 & 8   \\
(3456, 2160) & 7.46 & 14.18 & 88.00 & 50  \\
(3456, 2234) & 7.72 & 14.17 & 88.82 & 161 \\
\cmidrule(lr){2-5}
\textbf{Average} & \textbf{6.93} & \textbf{13.27} & \textbf{89.70} & \\
\midrule
\multicolumn{5}{l}{\textit{\textbf{Ultra-wide}}} \\
\midrule
(3840, 1080) & 4.15 & 11.93 & 87.85 & 181 \\
(5120, 1440) & 7.37 & 17.30 & 93.44 & 61  \\
\cmidrule(lr){2-5}
\textbf{Average} & \textbf{4.96} & \textbf{13.28} & \textbf{89.26} &  \\
\midrule
\multicolumn{5}{l}{\textit{\textbf{Extreme (4K+)}}} \\
\midrule
(3840, 2160) & 8.29  & 17.54 & 79.26 & 270 \\
(5120, 2880) & 14.75 & 17.51 & 82.86 & 70  \\
(6016, 3384) & 20.36 & 26.09 & 81.82 & 11  \\
\cmidrule(lr){2-5}
\textbf{Average} & \textbf{9.96} & \textbf{17.80} & \textbf{80.06} &  \\
\bottomrule
\end{tabular}
\end{table}

\section{Advanced Analysis of Agent Capabilities}
\label{sec:advanced}

\subsection{Analysis of Refusal Capability}
\label{app:refusal_agent}
This appendix elucidates the Refusal Agent's design, experimental methodology, prompt engineering, and exhaustive performance metrics, as introduced in the main paper. By isolating the feasibility assessment, we systematically analyze and elevate a VLM's capacity to reject infeasible instructions on the OSWorld-G benchmark. This detailed analysis supports the configuration chosen in our primary experiments, which aims to mitigate risks inherent in monolithic agents by balancing refusal accuracy with user experience.

\subsection{Experimental Protocol}
\begin{itemize}
    \item \textbf{Model:} Gemini-2.5-Pro
    \item \textbf{Task:} Binary feasibility judgment ("Yes" for executable, "No" for infeasible), parsed from a JSON output.
    \item \textbf{Dataset:} OSWorld-G, bifurcated into:
    \begin{itemize}
        \item \textbf{Refusal Subset:} 54 infeasible instructions (e.g., interacting with absent or non-existent UI elements).
        \item \textbf{Non-Refusal Subset:} 510 feasible instructions.
    \end{itemize}
    \item \textbf{Methodology:} We ablated three distinct prompt variants and two image scaling conditions (raw vs. 2$\times$/3$\times$ upscaling). Evaluations were conducted on the full subsets to ensure statistical fidelity, with results averaged over multiple runs for consistency.
\end{itemize}

\subsection{Prompt Formulations}
The following prompts were used to evaluate the Refusal Agent's performance under different reasoning conditions.

\begin{listing}[tb]
\caption{Basic Prompt for Feasibility Judgment (V0)}
\label{lst:prompt_basic_refusal}
\begin{lstlisting}
{
  "description": "Basic prompt for feasibility judgment without explicit reasoning",
  "content": "You are a model that can effectively judge whether to approve or reject tasks based on visual data. You should answer \"yes\" or \"no\" whether the given screen has the proper spot to execute the given user instruction. You should return only json format with embrace ```json``` without any comments.\n## example\n```json\n{  \n   \"answer\": \"#your answer\"\n}    ```"
}
\end{lstlisting}
\end{listing}

\begin{listing}[tb]
\caption{Reasoning Prompt (V1)}
\label{lst:prompt_reasoning_v1}
\begin{lstlisting}
{
  "description": "Reasoning prompt requiring step-by-step reasoning",
  "content": "You are a model that can effectively judge whether to approve or reject tasks based on visual data. You should answer \"yes\" or \"no\" from reasoning step whether the given screen has the proper spot to execute the given user instruction. You should return only json format with embrace ```json``` without any comments.\n## example\n```json\n{  \n   \"reasoning\": \"#your reasoning\",\n   \"answer\": \"#your answer from reasoning\"\n}    ```"
}
\end{lstlisting}
\end{listing}

\begin{listing}[tb]
\caption{Advanced Reasoning Prompt with Rules (V2)}
\label{lst:prompt_reasoning_v2}
\begin{lstlisting}
{
  "description": "Advanced reasoning prompt with explicit rules",
  "content": "You are an judge expert to predict whether the next gui action on the given image can excute or not. So You must analyze whether you have an area to execute user instruction from the given screen.\n- given rules to process\n    You must state the reason for your judgment in the reasoning field.\n    You must answer 'yes' or 'no' considering reasoning field.\n- given rules to judge\n    If there is even tiny information indicating that user instruction can be executed from the given screen, you must answer 'yes'.\n    If logically user instruction can't be executed from the given screen, you must answer 'no'.\n- given rules to output format\n    You should return only json format with embrace ```json``` without any comments.\n    ## format example\n```json\n{  \n   \"reasoning\": \"#your judgement\",\n   \"answer\": \"#your answer from reasoning\"\n}    ```"
}
\end{lstlisting}
\end{listing}

\subsection{Comprehensive Performance Analysis}
Our Refusal Agent significantly outperforms the baseline models reported in the original OSWorld-G paper. The baselines exhibit a wide spectrum of performance: Gemini-2.5-Pro (single-shot) serves as the \textbf{strongest baseline} but still only achieves \textbf{38.9\%} refusal accuracy, while other models show \textbf{moderate performance} (Seed1.5-VL at 18.5\%), \textbf{low refusal capability} (JEDI-7B at 7.4\%), or perform \textbf{no refusal detection} at all (0.0\%).

For our methods, we focused on balancing two critical metrics: \textbf{Refusal Accuracy} (correctly identifying infeasible instructions from the refusal subset) and the \textbf{False Positive Rate} (FPR), which measures the frequency of incorrectly rejecting feasible instructions from the non-refusal subset. The "Basic" prompt relies on direct judgment, while "Reasoning" prompts (V1 and V2) require the model to generate a rationale, which we found improves nuance. The goal is to maximize refusal accuracy while keeping the FPR low to ensure the agent remains useful and does not frustrate the user by rejecting valid commands. Image scaling enhances detection, with 2$\times$ scaling providing a significant boost and 3$\times$ scaling offering only marginal gains thereafter. The table below presents the quantitative results of this analysis.

\begin{table}[h!]
\centering
\caption{Refusal accuracy and false positive rate (FPR) on OSWorld-G. Refusal accuracy is measured on the infeasible subset (N=54), while FPR is derived from performance on the feasible subset (N=510).}
\label{tab:full_comparison_refusal_updated}
\small
\setlength{\tabcolsep}{5pt}
\begin{tabular}{@{} l c c @{}}
\toprule
\textbf{Model / Method} & \textbf{Refusal Acc. (\%)} & \textbf{FPR (\%)} \\
\midrule
\multicolumn{3}{@{}l}{\textit{\textbf{Baseline Models (Reported)}}} \\
Gemini-2.5-Pro & 38.9 & N/A \\
Seed1.5-VL & 18.5 & N/A \\
JEDI-7B/3B/OS-Atlas-7B & 7.4 & N/A \\
UGround/Aguvis/etc. & 0.0 & N/A \\
\midrule
\multicolumn{3}{@{}l}{\textit{\textbf{Our Methods (Gemini-2.5-Pro)}}} \\
Basic (Prompt V0) & 68.5 & 6.3 \\
Reasoning (Prompt V1) & 72.2 & 9.9 \\
Advanced Reasoning (V2) & 68.5 & \textbf{3.3} \\
Reasoning (V1) + 2$\times$ Scaling & \textbf{78.0} & 5.3 \\
Reasoning (V1) + 3$\times$ Scaling & 74.1 & 6.9 \\
\bottomrule
\end{tabular}
\end{table}

\paragraph{Discussion.}
The results in Table~\ref{tab:full_comparison_refusal_updated} illustrate a clear trade-off between maximizing refusal accuracy and minimizing false positives. The \textbf{Reasoning (V1) + 2$\times$ Scaling} method achieves the highest refusal accuracy (\textbf{78.0\%}), making it the most effective at identifying impossible tasks, which is critical for safety-focused applications. However, the \textbf{Advanced Reasoning (V2)} method achieves the lowest false positive rate (\textbf{3.3\%}) while maintaining a strong refusal accuracy of 68.5\%. This makes it the most reliable from a user-experience perspective, as it is the least likely to incorrectly reject a valid instruction.

This trade-off is critical: an agent that refuses too often (a high FPR) can be as problematic as one that never refuses. For the main results presented in this paper, we prioritized a smooth and reliable user experience, and therefore selected the \textbf{Advanced Reasoning (V2)} configuration. The choice of the optimal method ultimately depends on the application's priority: maximizing safety by catching all infeasible instructions, or ensuring robust, non-frustrating operation by avoiding incorrect rejections.

\subsection{Qualitative Failure Analysis}
\label{sec:qualitative_failures}

While MEGA-GUI sets a new state of the art in quantitative performance, a deeper qualitative analysis of its failure cases is essential for understanding the remaining challenges and guiding future research. This inspection reveals that errors are not random but fall into distinct archetypes, each highlighting the limitations of a core component within our multi-stage framework: the search policy, the grounding model, and the inter-agent communication pipeline.

\paragraph{Failure of Search: Attentional Fixation in the Zoom Policy.}
The first archetype is a failure of our core search mechanism, the Bidirectional ROI Zoom agent. On visually dense UIs with many repetitive, similar-looking elements (e.g., a layers panel in professional design software), the Stage 1 VLM (Gemini 2.5 Pro) can develop a persistent fixation on a salient but incorrect region. This triggers a non-productive loop where the agent confidently provides the same wrong coordinates. While our bidirectional policy is designed to recover by zooming out, a series of high-confidence errors can exhaust the error budget ($E_{\max}$) before the corrective mechanism can escape the loop, leading to premature termination. This demonstrates that even a robust search policy can be defeated by a model's powerful intrinsic biases.

\paragraph{Failure of Grounding: Deterministic Bias in the Grounding Agent.}
This second archetype highlights the limitations of the Stage 2 grounding VLM (UI-TARS-72B), illustrating the "No Free Lunch" principle. Failures here are not random but deterministic, rooted in the model's specific training data and amplified at low sampling `temperature`. For instance, if a model has been heavily fine-tuned on UIs that use a three-dots menu for "settings," it will invariably and confidently select that icon, even if the user's instruction and the current application's design clearly call for a gear icon. This occurs because the model's learned prior is so strong that it overrides the contextual information provided by the prompt. This reveals a critical challenge: a model's performance is path-dependent on its training history, creating predictable blind spots that our framework can mitigate but not always eliminate.

\paragraph{Failure by Instructional Over-Correction.}
The third archetype is a paradoxical failure mode where our Context-Aware Rewrite Agent, in its attempt to add precision, makes an instruction more difficult for the downstream grounding model. This occurs due to a "model dependency" mismatch: an instruction that is clearer for a human or a generalist LLM like GPT-4o may be counter-productive for a specialized grounding VLM like UI-TARS-72B. For example, given the simple command “\textit{Save the document},” the grounding model might correctly identify a text-free floppy disk icon based on its strong, learned visual association. However, the Rewrite Agent might "improve" the prompt to “\textit{Click the button labeled 'Save' next to the file name.}” The grounding model can now over-focus on the literal text "Save," fail to find it, and report an error, whereas the original, simpler instruction would have succeeded. This "curse of specificity" demonstrates that the interface between agents is critical; the optimal prompt is not universally explicit but is instead tightly coupled to the expectations and biases of the model that will execute it.

These three failure archetypes—spanning policy, model, and system levels—collectively demonstrate that the path to truly robust GUI automation involves more than simply improving individual model capabilities. They underscore the critical need for more adaptive policies that can detect non-productive loops, grounding models with less brittle conceptual understanding, and a more sophisticated orchestration strategy that tunes inter-agent communication. Addressing these system-level challenges represents a key direction for future work.

\section{Expanded Discussion on Broader Impacts}

The development of highly capable GUI agents like MEGA-GUI presents significant opportunities and underscores important areas for future research. A nuanced understanding of the technology's potential is essential for guiding responsible innovation.

\paragraph{Transformative Potential}
\begin{itemize}
    \item \textbf{Digital Accessibility:} The framework has the potential to dramatically improve digital accessibility. For users with motor impairments, the ability to operate complex, professional software like Figma or Adobe Photoshop using precise voice commands could be transformative, enabling new avenues for creative expression and employment.
    \item \textbf{Intelligent Automation:} In enterprise settings, MEGA-GUI could power the next generation of Robotic Process Automation (RPA). By automating tedious, repetitive workflows that currently require human interaction with graphical interfaces, it could improve efficiency, reduce human error, and free up employees for higher-value, strategic tasks.
\end{itemize}

\paragraph{Limitations and Future Directions}
While MEGA-GUI sets a new standard for performance, its reliance on multiple, sequential VLM calls introduces latency, which may be a constraint in real-time applications. Future work could explore model distillation to create smaller, more efficient specialist agents or investigate parallel processing strategies. Furthermore, the framework's performance is tied to the capabilities of proprietary models; continued research into high-performing open-source alternatives is crucial for broader accessibility and deployment.

\paragraph{Responsible AI Development}
The capabilities of advanced GUI agents also highlight the importance of responsible development. The architectural components of MEGA-GUI offer a strong foundation for building safe systems. The Refuser Agent, for example, serves as a critical guardrail against invalid instructions. A key direction for future research is to expand its capabilities beyond simple feasibility checks to include more sophisticated ethical reasoning, such as identifying and blocking instructions that could be manipulative, harmful, or violate an application's terms of service. Our decision to open-source the Grounding Benchmark Toolkit (GBT) is a deliberate step toward fostering transparent and collaborative research. By providing the community with standardized tools for evaluation, we aim to accelerate research into both the capabilities and the potential risks of such agents, empowering the community to develop effective safeguards and best practices collectively.

\end{document}